\documentclass{article}

    \PassOptionsToPackage{numbers, compress}{natbib}

\usepackage[preprint]{neurips_2025}

\usepackage[utf8]{inputenc} 
\usepackage[T1]{fontenc}    
\usepackage{hyperref}       
\usepackage{url}            
\usepackage{booktabs}       
\usepackage{amsfonts}       
\usepackage{nicefrac}       
\usepackage{microtype}      
\usepackage{xcolor}         

\usepackage{amsmath}
\usepackage{amssymb}
\usepackage{amsthm} 
\usepackage{amsfonts}
\usepackage{thmtools}
\usepackage{mathtools}
\usepackage{microtype}
\usepackage{graphicx}
\usepackage{booktabs} 
\usepackage{multirow}  
\usepackage{subcaption}
\usepackage{adjustbox}
\usepackage{svg}
\usepackage{tikz}
\usepackage{authblk}

\usepackage{algorithm}
\usepackage{algpseudocode}

\usetikzlibrary{3d}

\newtheorem{proposition}{Proposition}

\title{ACE and Diverse Generalization via\\ Selective Disagreement}

\newcommand{\corresponding}{$^{*}$}
\newcommand{\cut}[1]{}
\newcommand{\mbf}[1]{\mathbf{#1}}

\DeclareMathOperator*{\argmax}{arg\,max}

\author[1]{Oliver Daniels \corresponding}
\author[4]{Stuart Armstrong \corresponding}
\author[2]{Alexandre Maranhão}
\author[3]{Mahirah Fairuz Rahman}
\author[1]{Benjamin M. Marlin}
\author[4]{Rebecca Gorman}
\affil[1]{University of Massachusetts Amherst}
\affil[2]{CentraleSup\'elec}
\affil[3]{University of Oxford}
\affil[4]{AlignedAI}
\affil[$*$]{\texttt{\{odanielskoch@umass.edu, stuart@buildaligned.ai\}}}

\begin{document}

\maketitle

\begin{abstract}
Deep neural networks are notoriously sensitive to spurious correlations - where a model learns a shortcut that fails out-of-distribution. Existing work on spurious correlations has often focused on incomplete correlations,leveraging access to labeled instances that break the correlation. But in cases where the spurious correlations are complete, the correct generalization is fundamentally \textit{underspecified}. To resolve this underspecification, we propose learning a set of concepts that are consistent with training data but make distinct predictions on a subset of novel unlabeled inputs. Using a self-training approach that encourages \textit{confident} and \textit{selective} disagreement, our method ACE matches or outperforms existing methods on a suite of complete-spurious correlation benchmarks, while remaining robust to incomplete spurious correlations. ACE is also more configurable than prior approaches, allowing for straight-forward encoding of prior knowledge and principled unsupervised model selection. In an early application to language-model alignment, we find that ACE achieves competitive performance on the measurement tampering detection benchmark \textit{without} access to untrusted measurements. While still subject to important limitations, ACE represents significant progress towards overcoming underspecification.

\end{abstract}

\section{Introduction}

Standard deep learning approaches often struggle to maintain high performance under distributional shifts \citep{hendrycks2019benchmarkingneuralnetworkrobustness}. Models are particularly sensitive to distributional shifts that ``break'' the correlation (present on the training distribution) between ``spurious'' and ``ground truth'' features (e.g., classifying huskies based on the presence of snow \citep{ribeiro2016whyitrustyou}, or images based on texture rather than shape \citep{geirhos2022imagenettrainedcnnsbiasedtexture}). 

In regards to spurious correlations, existing work largely focuses on \textit{incomplete} spurious correlations, relying on various forms of reweighting and up-sampling of labeled instances where the correlation breaks \citep{sagawa2020distributionallyrobustneuralnetworks, kirichenko2023layerretrainingsufficientrobustness, nam2020learningfailuretrainingdebiased, liu2021justtraintwiceimproving}. The present work treats the more challenging problem of \textit{complete} spurious correlation, where the ground-truth and spurious feature are perfectly correlated on a source distribution, but come apart on a target distribution. In such cases, the correct generalization is fundamentally \textit{underspecified} \citep{d2022underspecification}. 

Underspecification can arise in a range of settings, but is particularly relevant to outstanding problems in AI alignment (e.g. easy-to-hard generalization \citep{hase2024unreasonableeffectivenesseasytraining, burns2023weak} and goal (mis-)generalization \citep{langosco2023goalmisgeneralizationdeepreinforcement}) where distributional shifts cause perfectly correlated variables to come apart. Work on the ``science of deep learning'' and ``interpretability'' advocate a bottom up approach, attempting to rigorously understand and shape the learning process and representations of deep networks. While such methods have shown some initial promise
\citep{marks2024sparsefeaturecircuitsdiscovering,lehalleur2025eataialignment},
they are currently far from reliably characterizing out-of-distribution behavior. 

In this work, we propose a more exhaustive method: learning an ensemble of concepts 
consistent with the training data that make distinct predictions on novel unlabeled inputs.
To learn these \textit{diverse generalizations} of the labeled data, we introduce a loss function that encourages \textit{selective}, \textit{confident} disagreement among the ensemble members.  
Equipped with a diverse set of concepts, 
practitioners can either select the desired generalization 
with a small number of ground-truth labels from the target distribution, 
or use the entire ensemble to better calibrate uncertainty estimates. 
We call this general framework \textit{Algorithm for Concept Extrapolation}\footnote{patent pending} (ACE).

By encouraging \textit{confident} disagreement, ACE disentangles candidate generalizations.
By encouraging \textit{selective} disagreement, ACE promotes learning ``naturally'' diverse hypotheses that do not ``artificially'' disagree on correlation-aligned instances. 
Empirically, we find that across a range of complete-correlation image and language benchmarks ACE learns hypotheses on the target distributions that correspond to semantically meaningful human concepts. ACE also achieves strong performance 
in the measurement tampering detection task \citep{roger2023benchmarksdetectingmeasurementtampering}, 
a positive early sign for the applicability of ACE to scalable oversight \citep{amodei2016concreteproblemsaisafety, bowman2022measuring}.

Two concurrently developed methods, DivDis\citep{lee2023diversifydisambiguatelearningunderspecified} 
and D-BAT \citep{pagliardini2022dbat}, 
propose similar approaches, using disagreement losses that encourage 
statistical independence (DivDis) or maximal disagreement (D-BAT). 
However, these loss functions make implicit assumptions about how the concepts are correlated on the target distribution. DivDis works best when the concepts are uncorrelated (correlation of $0$) and D-BAT works best when the concepts are anti-correlated (correlation of $-1$ relative to their spurious correlation in the source distribution). Furthermore, these methods offer no principled model selection criteria for early stopping or hyperparameter tuning \citep{gulrajani2020searchlostdomaingeneralization}.
In contrast, ACE only assumes an explicit configurable \textit{lower bound} on the mix rate, that is, the probability that any two concepts disagree in the target distribution, and provides
effective methods for model selection and identifying mix rate without access to labeled target data. 

The remainder of this paper is organized as follows:
In \S\ref{section:background}, we formalize the problem of diverse generalization
In \S\ref{section:method}, we introduce and motivate the proposed approach. 
In \S\ref{section:experiments}, we present experiments and results. 
In \S\ref{section:conclusion}, we summarize our work and discuss limitations as well as future directions.
We cover related work in Appendix \ref{appendix:related-work}.

\begin{figure}[t]
    \begin{minipage}[c]{0.55\textwidth}
        \input{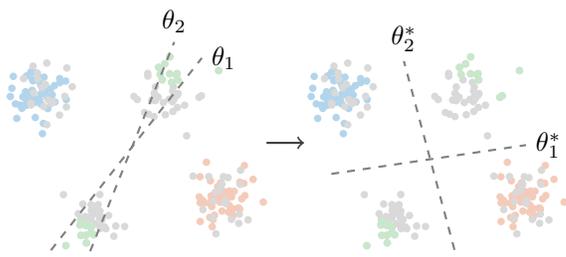}
    \end{minipage}%
    \begin{minipage}[c]{0.45\textwidth}
        \caption{\textbf{Confident Selective Disagreement} 
        On the left, a pair of decision boundaries $\theta_1, \theta_2$ initially trained to distinguish labeled blue and red
        points. Given additional unlabeled points (gray), ACE trains an ensemble to confidently disagree 
        on the top-k points (green) where the concepts are most likely 
        to disagree (according to the current ensemble).
        By increasing the margin from the green points, the 
        resulting decision boundaries $\theta_1^*, \theta_2^*$ avoid intersecting high density regions.}
        \label{fig:bootstrap_disagreement}
    \end{minipage}
\end{figure}

\section{Problem formulation}
\label{section:background}
\cut{***************
Humans use concepts to divide elements of the world into sets. We have the concept `dog breeds', for instance, with which we can map dogs into concepts corresponding to breeds: corgis, labradors, poodles, etc. Machine learning-based classification algorithms generate label functions, which map inputs to labels. In specific environments, these label functions can correspond to human concepts. The aim of machine learning for classifiers is to generate label functions that correspond to a desired concept, in as large a test environment as possible.
****************}

Let $\mathcal{X}$ be a space of possible inputs and $\mathcal{Y}$ be the label space. Let $C: \mathcal{X} \rightarrow \mathcal{Y}$ be a human concept mapping inputs to labels. Let $p_S(x, y)$ be a \textit{source} distribution of inputs $x \in \mathcal{X}$  and labels $y\in\mathcal{Y}$. Using $\Delta$ to designate probability distributions over a space, $p_S \in \Delta(\mathcal{X} \times \mathcal{Y})$. Assume that on the source distribution, the labels and the concept are perfectly correlated. For $\rho$ the Pearson correlation coefficient, $\rho_{p_S} (C(x), y) = 1$.
Assume we also have a target distribution $p_T(x) \in \Delta(\mathcal{X})$ which carries no label information.

Concept extrapolation is defined as the problem of producing a classifier $p_{\theta}(y|x)$ parameterized by $\theta \in \Theta$ that reproduces a desired concept $C$ on $p_T$ using only labeled instances from $p_S$ and unlabeled instances from $p_T$. 
A classifier $p_\theta:\mathcal{X}\to\Delta(\mathcal{Y})$ that solves the concept extrapolation problem should minimize the expected generalization error
\cut{
\footnote{A more sematic question is whether an algorithm that does concept extrapolation is learning the concept $C$, or just its values in $p_T$. Without taking a position on the philosophical question of learning, it does seem to us that if the algorithm successfully does concept extrapolation for all possible $p_T$, then it can be considered to have learned $C$ in practical terms.}
}
$\epsilon_T^C({\theta}) =\mathbb{E}_{ p_T}[C(x)\neq \argmax_y p_\theta(y|x)]$.
Because the target distribution is unlabeled, we cannot directly approximate the expected generalization error as required by ERM \citep{vapnik1999nature}. While we can approximate  the expected generalization on the source distribution, $\epsilon_S^C({\theta})$, this expectation can differ significantly from $\epsilon_T^C({\theta})$ \citep{ben2006analysis}. In the terminology of \citep{armstrong_model_splintering}, model splintering can occur: models designed exclusively for $p_S$ may no longer function as intended on $p_T$.

\cut{***
The concept extrapolation process is underspecified  in multiple ways \citep{d2022underspecification}. An algorithm may produce a classifier $p_{\theta}$ that fails to correspond to any human concept on $p_T$. However, it might also produce a classifier that does correspond to a human concept, but not to the intended one, $C$.
***}

A particularly concerning issue for concept extrapolation is complete spurious correlation of multiple concepts.  Let $\mathcal{C} = \{C_i\}_{i=1}^F$  denote a set of (human) concepts with $C=C_1$ the desired concept. We assume all concepts are completely correlated on the source distribution: $\rho_{p_S} (C_i(x),C_j(x))=1$ for each $C_i, C_j$ pair, but all disagree with some probability on the target distribution: $\rho_{p_T} (C_i(x),C_j(x))< 1$. 

Under complete correlation, identifying the intended concept $C$ is fundamentally underspecified \citep{d2022underspecification}. Instead, our paper addresses the problem of learning an ensemble of classifiers parameterized by $\boldsymbol{\theta} = \{\theta_i\}_{i=1}^{F}$ with each classifier $p_{\theta_i}(y|x)$ corresponding to a distinct correlated concept. Letting $\pi(i)=j$ represent the unknown true mapping between classifiers $p_{\theta_i}(y|x)$ and corresponding concepts $C_j$,  our goal is to minimize the following expected generalization error:

\begin{align}\label{target:div_gen_error}
    \epsilon_T^\mathcal{C}({\theta}) =\sum_{i=1}^F\mathbb{E}_{p_T}[ C_{\pi(i)}(x)\neq \argmax_y p_{\theta_i}(y|x)],
\end{align}

We refer to this problem as \textit{diverse generalization}. As an \textit{intermediate} goal to concept extrapolation, diverse generalization produces a set of candidate classifiers from which the classifier corresponding to the concept of interest $C$ can be identified with a small amount of labeled target data. In general, there may be a more complex relationship between the set of classifiers learned and the desired concept extrapolating classifier such that $\theta=f(\boldsymbol{\theta})$ for some ``extrapolation'' function $f$, but even in these cases we expect learning a ``basis'' for distinct concepts to be helpful. 

\subsection{Mix Rate}
When designing and evaluating diverse generalization algorithms, the frequency of disagreement between concepts is an important quantity to track. Let $\mathcal{G}' = \{[c_1, \ldots, c_F] \in \{0,1\}^F| \exists(c_i \neq c_j)\}$ denote the set of ``disagreement groups'': the combination of possible concept outputs where at least two concepts disagree, with $\mathbf{c}^g$ denoting the labeling of the $g$-th group. Then we define the mix rate as the expected frequency of which any two concepts disagree:

\begin{equation}
r = \mathbb{E}_{p_T}[(C_1(x) \dots C_F(x)) \in \mathcal{G}']
\end{equation}

We can similarly define group mix rates as the expected frequency of a particular group $r_g = \mathbb{E}_{p_T}[(C_1(x) \dots C_F(x))=\mathbf{c}^g]$. 

As an example, in the case $F=2$, the disagreement groups are $\mathbf{c}^1 = (0, 1)$ and $\mathbf{c}^2 = (0, 1)$, and the mix rate is the frequency at which they appear in the target distribution.

\section{Diverse Generalization with ACE}
\label{section:method}
ACE is a class of algorithms for concept extrapolation that attempt to find approximate solutions for diverse generalization, followed by a process for combing ensemble members to extrapolate the intended concept. ACE is motivated by the following insight: if correlated concepts disagree on some subset of the target distribution, then we can identify these concepts by training ensemble members to \textit{confidently} disagree on the subset of data where they \textit{already} disagree the most. In this way, ACE bootstraps default ensemble diversity \citep{lakshminarayanan2017simplescalablepredictiveuncertainty} into an ensemble with more distinct concepts. See Figure ~\ref{fig:bootstrap_disagreement} for an illustration of this bootstrapping dynamic.

In this paper, we focus on the specific case of binary image and text classifiers $\mathcal{Y} = \{0, 1\}$, using a top-K pseudo-label loss function on the target data to encourage confident and selective disagreement. 

\subsection{Loss Function}
\label{sec:topk-loss}
Given a labeled source dataset $\mathcal{D}_S = \{(x_i, y_i)\}_{i=1}^{N_S}$ and unlabeled target dataset $\mathcal{D}_T = \{x_i\}_{i=1}^{N_T}$ sampled from the source and target distributions, the objective function used to train the ensemble consists of corresponding loss functions $\mathcal{L} = \mathcal{L}_S + \mathcal{L}_T$. 
Because the concepts are assumed to be perfectly correlated with each other and the label $y$ on the source distribution, we apply a standard negative log-likelihood loss to both heads using the source data: $\mathcal{L}_S(\boldsymbol{\theta}, \mathcal{D}_S) = \frac{1}{N_S}\sum_{i=1}^{N_S}\sum_{j=1}^{F}\ell(y_i|x_i, \theta_j)$ with $\ell$ the negative log-likelihood. 

On the target distribution, we apply a pseudo-labeling procedure that encourages selective confident disagreement. First, for each disagreement group $g \in \mathcal{G}'$, we compute the probability that each instance belongs to $g$ ($p_{\boldsymbol{\theta}}^g(x_i)=\prod_{j=1}^F p_{\theta_j}(c^g_j|x_i)$), yielding the set $\mathcal{P}^g(\boldsymbol{\theta}) = \{p_{\boldsymbol{\theta}}^g(x_i)\}_{i=1}^{N_T}$. Next, on the assumption that \textit{some fraction} of the target instances belong to each disagreement group at a rates lower bounded by $\bar{r}_g$ (such that there at least $k_g = \lfloor N_T \cdot \bar{r}_g\rfloor$ instances belong to the group in the dataset), for each group we take the negative log likelihood loss with respect to the group concept labels $\mathbf{c}^g$ on the top-k instances most likely to belong to each group. Computationally, we define group masks $\mathbf{m}_g$ with $m_{i,g} = 1 \text{ if  } i \in \text{topk}(k_g, \mathcal{P}^g(\boldsymbol{\theta})) \text{ else } 0$, yielding the loss function:
\begin{align}
\mathcal{L}_T(\theta,\mathcal{D}_T,\mbf{m}) &= 
\sum\limits_{g=1}^{|\mathcal{G}'|} \frac{\lambda_g}{\mathbf{1}^{\top}\mathbf{m}_g}\sum\limits_{i=1}^{N_T}\sum\limits_{j=1}^F 
m_{i,g} \cdot \ell(c^g_j|x_i, \theta_j)
\label{eq:ace_loss}
\end{align}
Because the top-k function is non-differentiable, we apply a coordinate-wise approach, iteratively updating the mask variables and the parameters: 
\begin{align}
    m_{i,g} &\leftarrow 
    \begin{cases}
        1 & \mbox{ ... if } i\in \mbox{topk}(k_g,\mathcal{P}^g(\boldsymbol{\theta}^{t})) \\
        0 & \mbox{ ... otherwise}
    \end{cases}\\
    \boldsymbol{\theta}^{t+1} &\leftarrow \boldsymbol{\theta}^{t} - \alpha\cdot \nabla_{\boldsymbol{\theta}} \mathcal{L}(\boldsymbol{\theta}^{t},\mathcal{D}_S,\mathcal{D}_T,\mbf{m})
\end{align}




\subsection{Intuition}
To illustrate how Equation~\ref{eq:ace_loss} encourages selective disagreement, we present the loss in the case of $F=2$. Letting $\text{TopK}(\mathbf{c}^g) = \text{topk}(k_g, \mathcal{P}^g(\boldsymbol{\theta}))$ denote set of indices for the top-k most likely instances to belong to each group (and dropping the regularization weights) we have:
\begin{align*}
\mathcal{L}_T(\theta,\mathcal{D}_T) =\; &
\frac{1}{k_{(0, 1)}} 
\sum_{i \in \text{TopK}{(0,1)}} \left[
\ell(0 \mid x_i, \theta_1) + \ell(1 \mid x_i, \theta_2)
\right] \\
+\, &
\frac{1}{k_{(1, 0)}} 
\sum_{j \in \text{TopK}{(1,0)}} \left[
\ell(1 \mid x_j, \theta_1) + \ell(0 \mid x_j, \theta_2)
\right]
\end{align*}

This loss will encourage the two heads to disagree more confidently on the instances where they already disagree the most. For example, for the $(0, 1)$ group,  head one will be more confident in predicting 0 and head two will be more confident in predicting 1, and vice-versa for instances in $(1, 0)$.

\subsection{Implementation Details}
We now describe a number of implementation details. First, we apply \textit{stochastic} gradient descent and update $\mathbf{m}$ and $\boldsymbol{\theta}$ at every \textit{batch}. Second, inspired by incremental thresholds used in pseudo-labeling \citep{Lee2013PseudoLabelT}, we apply a linear schedule to the mix rate lower bound, incrementing from $0$ to $\bar{r}$ over $\mathcal{T}$ iterations, with $\bar{r}_t = \min\left(\frac{t}{\mathcal{T}} \cdot \bar{r}, \bar{r}\right)$. In practice we update $\bar{r}$ after every training \textit{epoch}. Third, by default we assume balanced group distributions, such that given a global mix rate lower bound $\bar{r}$ group mix rate lower bounds are defined as $\bar{r}_g = \bar{r}/|\mathcal{G}'|$. To relax the assumption of evenly balanced groups, we can instead aggregate the top-k loss  across groups by selecting the $k = \lceil N_T \cdot \bar{r} \rceil$ (instance, group) pairs with the highest probability of disagreement. Letting  $\mathcal{P}(\theta) = \{p_{\boldsymbol{\theta}}^g(x_i) \text{ for } i=1\dots N_T, g=1\ldots |\mathcal{G}'|\}$ be the combined set of group probabilities for each instance, we define a new function $\mbox{topk}(k,\mathcal{P}(\theta))$ to return the set of $(i,g)$ index tuples of elements in $\mathcal{P}(\theta)$ that have the $k$ largest values and set the mask accordingly. We refer to this variant as the \textit{group aggregated} loss. 

As a model selection criteria for hyperparameter tuning and early stopping, we compute total validation loss on held-out subsets of source and target distributions \citep{albuqerque2019}. The validation loss holds $\lambda_g =1$ and the target mix rate lower bound $\bar{r}$ fixed. Notably, unlike prior work, this method does not use oracle test data for tuning. See Appendix~\ref{appendix:model_selection} for details on our model selection approach and why our proposed loss is better suited for model selection than prior methods.

\subsection{Motivation}
\label{sec:motivation}
We describe three key benefits of ACE over prior diverse generalization methods.

\textbf{Low Density Separation}: By training against hardened pseudo-labels, ACE implicitly implements an entropy penalty, pushing decision boundaries towards \textit{low density regions} of representation space \citep{grandvalet2004semi, Lee2013PseudoLabelT} - see Figure~\ref{fig:bootstrap_disagreement}. Assuming (human) concepts also lie in low density regions (appealing to the natural abstractions hypothesis \citep{Wentworth2020Alignment, huh2024platonicrepresentationhypothesis}), this promotes learning \textit{disentangled} concepts rather than concept interpolations. In contrast, DivDis \citep{lee2023diversifydisambiguatelearningunderspecified} provides no entropy regularization, and suffers relatively poor combined concept accuracy on real-world datasets (see Appendix~\ref{appendix:mix-rate-results}). 

\textbf{Stable Joint Training} By only taking gradients against hardened pseudo-labels, ACE is stable under joint training of heads, avoiding the need to iteratively train each ensemble member against the output of the others. Such iterative training can be problematic when there is not a strong bias towards a single concept, as the default decision boundary maximizes margin on the source distribution by intersecting the disagreement groups in representation space. For illustration of this dynamic, see the initial decision boundaries in ~\ref{fig:bootstrap_disagreement}. In contrast, D-BAT \citep{pagliardini2022dbat} is unstable under joint training and requires iterative fitting, and as a result  suffers poor performance on datasets without a strong simplicity bias (see Appendix~\ref{appendix:mix-rate-results}).

\textbf{Proper Scoring} Loss functions
should ideally assign correct hypotheses a globally minimal loss. In the language of proper scoring \citep{gneiting2007strictly}, a scoring rule for a distribution $p$ that approximates $q$ is proper if $S(p, q) \geq S(q,q)$. Under perfect spurious correlation, ACE is a proper scoring rule if mix rate lower bounds $\bar{r}_g \leq r_g $ hold. In contrast,  DivDis \citep{lee2023diversifydisambiguatelearningunderspecified}
and D-BAT, \citep{pagliardini2022dbat} are only proper  when target distribution concepts have 0 mutual information (typically $r=0.5$), or are perfectly 
anti-correlated ($r=1$). See Appendix~\ref{appendix:divdis_mix_rate_proof} for proper scoring proofs for all three loss functions, and Section~\ref{sec:mix_rate_robustness} for empirical 
results showing both D-BAT and DivDis are sensitive to changes in the mix rate that violate their proper scoring requirements.

\section{Experiments}
\label{section:experiments}
Here we aim to test the following research questions:
1) How well does ACE learn concept-aligned hypotheses from completely correlated data?
2) How sensitive are ACE and existing methods are to variation in 
the mix rate?
3) How well does ACE performs under \textit{incomplete} spurious correlations?
4) Can we infer mix rates from changes in (semi-supervised) validation loss?
We also provide additional experiments on the efficacy of model selection strategies, and whether 
alternative concept labels improve performance. 
See Appendix~\ref{appendix:cc-datasets} for a description of the datasets, Appendix~\ref{appendix:experiment_details} for details on the experimental setup and a list of hyperparameters, and Appendix~\ref{appendix:model_selection} for details on our model selection approach.

\begin{figure}[ht]
    \includegraphics[width=0.9\textwidth]{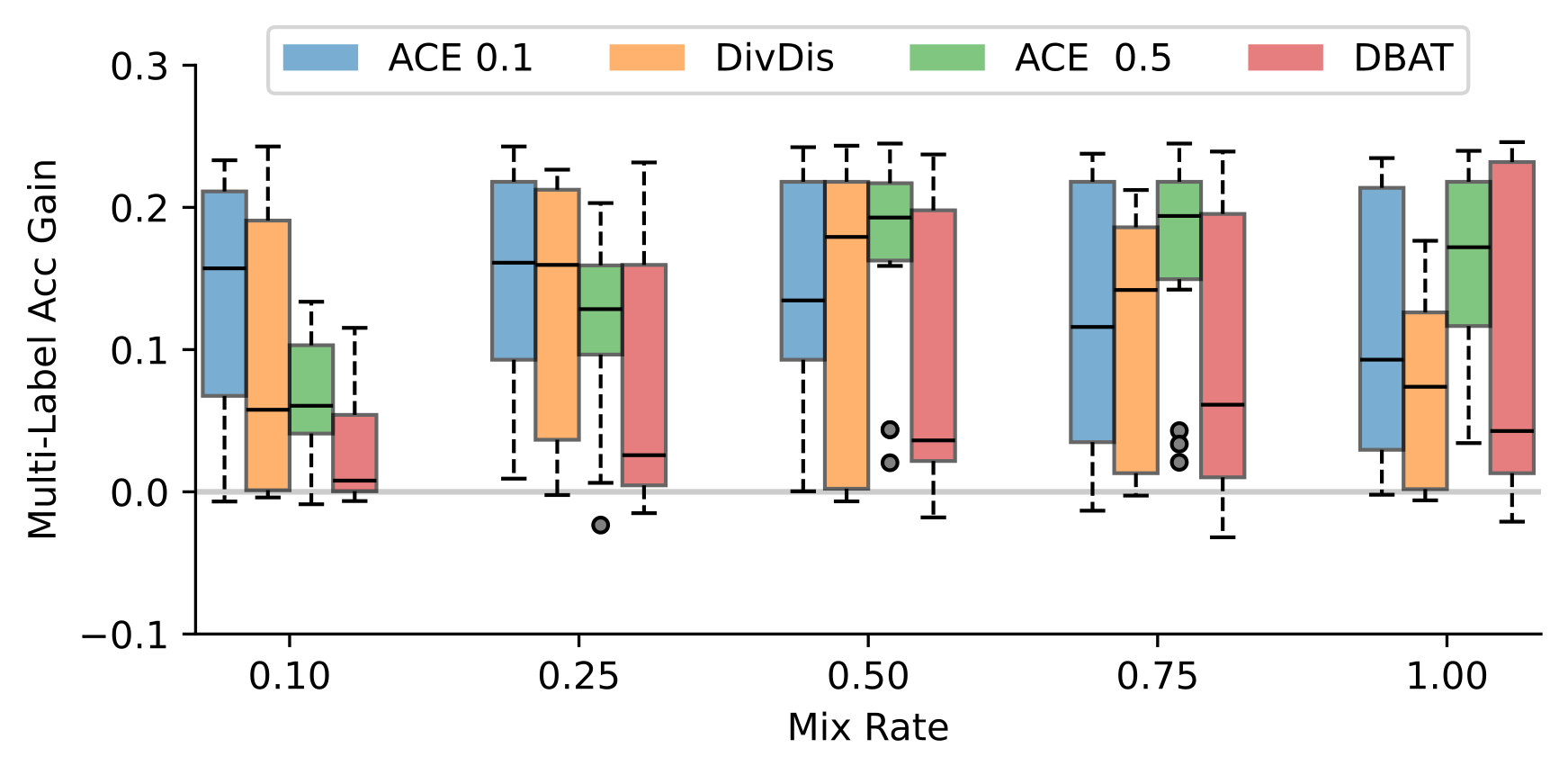}
    \caption{\textbf{Aggregated Multi-Label Accuracy Across Mix Rates} 
        Box plots of multi-label (ground truth and spurious) accuracy 
        gain over ERM of diverse generalization methods across mix rates, 
        averaged over 6 datasets with complete spurious correlations 
        and 3 random seeds for each configuration.
        Heads are assigned to labels using oracle test data.
        ACE 0.1 outperforms all other methods at a mix rate of 0.1, 
        but suffers an (absolute) drop in performance as mix rate increases. 
        ACE 0.5 out performs all other methods on mix rates at or above 0.5, 
        but performs poorly at mix rates below the assumed lower bound. 
        DivDis performs best at a mix rate of 0.5, 
        D-BAT performs best at a mix rate of 1.0, and performance of both methods decays as the mix rate deviates from these values. 
        D-BAT has low median accuracy across mix rates, likely due to datasets with small or no difference in the simplicity of ground-truth and spurious concepts (see Section~\ref{sec:motivation}).
    }
    \label{fig:cc_mix_rate_avg_acc_agg}
\end{figure}

\subsection{Robustness to Mix Rate Variations}
\label{sec:mix_rate_robustness}
To test the sensitivity of ACE and existing methods to variations in the 
target distribution mix rate, we evaluate ACE, DivDis, D-BAT, and an ERM baseline 
on image and language datasets
with \textit{complete} spurious correlations between two clear human concepts $C$ (the desired concept) and $C'$ (another concept, generally of lower simplicity on the source dataset).
We artificially varied the target mix rates $r$ to have values $[0.1, 0.25, 0.5, 0.75, 1.0]$, preserving the relative correlation-conflicting group frequencies. We experiment with a low and high mix rate setting for ACE loss,  
with mix rate lower bounds of $0.1$ and $0.5$. 

For the three simple balanced datasets (Toy Grid, FMNIST-MNIST, and CIFAR-MNIST), 
we evenly distribute the mix rate lower bound across both minority groups. 
For the more realistic imbalanced datasets (Waterbirds-CC, CelebA-CC, Multi-NLI-CC-Binary), 
we use the group aggregated loss (see Section~\ref{sec:topk-loss}). See Appendix~\ref{appendix:comb-group-loss-ablation} for 
ablation results using the opposite configuration. 
Aggregated results for all methods across all datasets and mix rates are shown 
in Figure~\ref{fig:cc_mix_rate_avg_acc_agg}. Overall, the results are consistent with our 
hypotheses: ACE outperforms prior methods when the mix rate lower bound holds and is relatively tight, and loss functions from prior methods are sensitive to variations in mix rate that violate proper scoring. See Appendix~\ref{appendix:mix-rate-results} for disaggregated results on each dataset.

\subsection{Incomplete Spurious Correlations}
We are also interested in how ACE 
handles incomplete spurious correlations - where the source distribution contains 
correlation-conflicting instances. \footnote{
Incomplete correlations can also be seen as label bias from the perceptive of the alternate concept $a_s \neq a_y$.
}
To study this, we evaluate ACE, D-BAT, and DivDis on (an unmodified version of) Waterbirds

We experiment with the normal setting of the algorithm (using the same (bird) label $y$ for both classifiers on the source distribution), and an additional setting where we assume ground-truth access to an alternate (background) label $y'$.  
We use all the same hyperparameters from the previous section, but do not use mix rate 
scheduling. We use the same model selection criteria as before (validation loss), 
except for D-BAT, where we use validation source accuracy 
\footnote{
    Using validation loss for D-BAT model selection resulted in 
    extremely poor performance, likely because D-BAT is able to ``overfit'' to the disagreement loss. 
    We return to the idea of using validation source accuracy for selecting degree of disagreement  
    in Section~\ref{sec:mix_rate_lower_bound_selection}.
}.

The results of the experiments are shown in Table~\ref{tab:incomplete_spurious_correlation_results}. 
We find that ACE 0.5 achieves comparable ground-truth accuracy to DivDis, and substantially 
higher spurious attribute and worst-group accuracy. We also find (unsurprisingly) that 
group labels substantially improve spurious concept accuracy. 
Group labels also dramatically improve the \textit{target} concept accuracy of D-BAT, likely because the 
first head learns a ``cleaner'' concept for background that more strongly disagrees with bird 
on the target distribution (see discussion in Section~\ref{sec:motivation}).

\begin{table*}
\centering
\begin{tabular}{lcccccc}
\toprule
& \multicolumn{3}{c}{Single Label ($y$, $y$)} & \multicolumn{3}{c}{Concept Labels ($y$, $y'$)} \\
\cmidrule(lr){2-4} \cmidrule(lr){5-7}
Method & Avg $y$ & Avg $y'$ & Worst $y$ & Avg $y$ & Avg $y'$ & Worst $y$ \\
\midrule
ACE 0.1 & 88.9 ± 1.3 & 74.0 ± 3.6 & 53.9 ± 12.8 & 87.9 ± 0.7 & 89.7 ± 2.6 & 54.3 ± 8.8 \\
DivDis & \textbf{91.0 ± 3.2} & 72.9 ± 1.3 & \textbf{67.7 ± 8.1} & \textbf{91.7 ± 1.4} & 92.4 ± 0.9 & 66.9 ± 3.2 \\
ACE 0.5 & \textbf{92.8 ± 0.4} & \textbf{90.7 ± 1.1} & \textbf{70.9 ± 2.3} & \textbf{92.5 ± 0.9} & \textbf{94.8 ± 0.5} & \textbf{75.6 ± 7.0} \\
D-BAT & 60.2 ± 19.4 & 68.7 ± 1.5 & 33.1 ± 23.4 & 89.6 ± 0.8 & 91.8 ± 0.7 & 50.3 ± 5.4 \\
ERM & 84.7 ± 2.9 & 63.1 ± 2.9 & 51.1 ± 10.2 \\
\bottomrule
\end{tabular}
\caption{
    \textbf{Incomplete Spurious Correlation Accuracy} 
    Test accuracy on Waterbirds with \textit{incomplete} spurious correlations 
    using single label ($y=C_{\text{bird}}$) and group labels ($y=C_{\text{bird}}$, $y'=C_{\text{back}}$) on the source distribution. Oracle test data is used to 
    With the single label, DivDis and ACE 0.5 achieve comparable average and worst-group accuracy, 
    but ACE 0.5 achieves substantially higher attribute accuracy. 
    Using separate concept labels substantially improves the performance of D-BAT.  
    Training with concept labels also improves background accuracy for all other methods, 
    though ACE 0.5 still achieves the highest background accuracy. 
}
\label{tab:incomplete_spurious_correlation_results}
\end{table*}

\subsection{Mix Rate Lower Bound Selection}
\label{sec:mix_rate_lower_bound_selection}
Thus far we have assumed fixed mix rate lower bounds 0.1 and 0.5 corresponding to 
``low'' and ``medium'' mix rate regimes. While ACE allows practitioners to set mix rate lower bounds 
based on domain knowledge, we would like a procedure for \textit{inferring} the mix rate lower bound 
from the target distribution. 

As one approach to mix rate inference, we propose tracking validation loss across a range of $\hat{r}$ values. Typically, semi-supervised losses would 
be hard to compare across different configurations, as changing the hyperparameters changes the underlying quantity being measured. 
But for ACE, since we take the \textit{mean} validation loss for each group, in theory the validation 
losses should be similar across mix rate lower bounds until the mix rate lower bound exceeds 
the ground truth mix rate. Under this assumption, we can run ACE with a range of mix rate lower bounds, selecting the \textit{highest} lower bound that does not result in a substantial 
drop in validation loss. 

\begin{figure}
    \centering
    \includegraphics[width=\columnwidth]{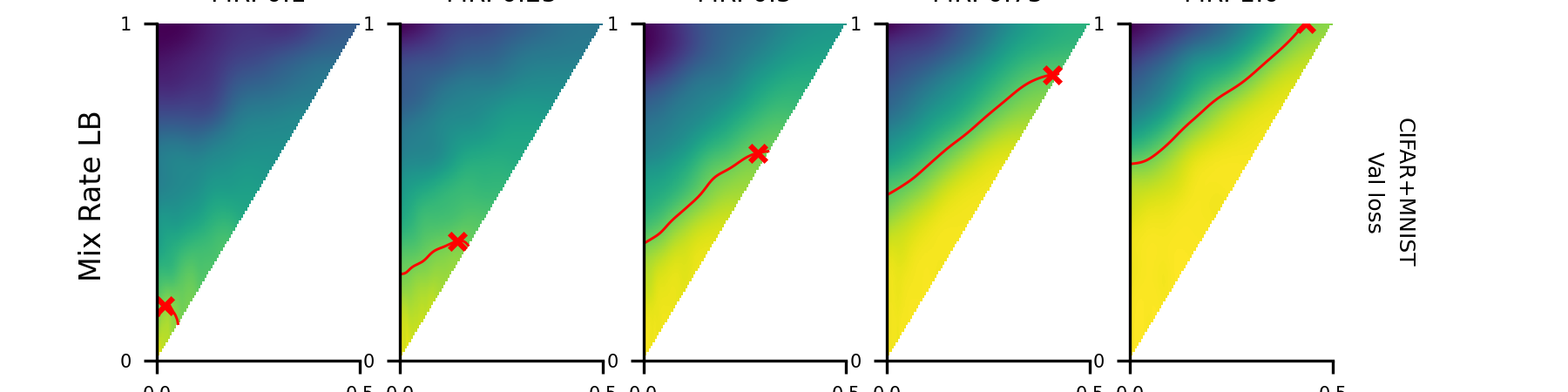}
    \caption{\textbf{Validation Loss Density Estimate for CIFAR-MNIST} 
        Validation loss density estimate for CIFAR-MNIST as a function of mix rate lower bound 
        and smallest group lower bound. Selecting the highest lower bound subject to 
        being within the 20th percentile of the validation loss density estimate provides a good signal 
        for mix rate lower bound. 
    }
    \label{fig:val_loss_cifar_mnist_density_grid_main}
\end{figure}

We evaluate this approach on our four image datasets. Concretely, we run ACE across a triangular grid, varying the minimum group mix rate and total mix rate, then construct a gaussian density estimate of validation loss. 
See Figure~\ref{fig:val_loss_cifar_mnist_density_grid_main} for the validation loss density estimate 
for CIFAR-MNIST, and Appendix~\ref{appendix:val_loss_density_estimates} for the validation loss 
and test accuracy density estimates for each of the four datasets. 
We find validation loss provides a 
very strong signal for mix rate on the synthetic datasets and 
moderately strong signal on the real-world datasets. 
For a selection heuristic, picking the highest lower bound within 20th percentile validation loss density behaves reasonably well, 
but this threshold was chosen in an ad-hoc manor using oracle information about the mix rate and test performance. Future work should evaluate how well thresholds generalize to other datasets and models. 

\section{Application: Measurement Tampering Detection}
As a case study for the applicability of concept extrapolation to scalable oversight \citep{bowman2022measuring}, we evaluate ACE as a method for measurement tampering detection (MTD) \citep{roger2023benchmarksdetectingmeasurementtampering}. In MTD, a prediction model is trained on multiple measurements of agent behavior. Measurements are sampled from the trusted distribution (where the measurements are always accurate) and an untrusted hard distribution (where agents may \textit{tamper} with measurements to hide bad outcomes). MTD methods must use this prediction model to distinguish ``real'' and ``fake'' positives on the untrusted distribution, and are evaluated according to  AUROC on an evenly balanced held-out test set. Mapping to our framework, $C_{y}$ is the ground-truth intended concept, $C_m$ is the conjunction of all the measurements \footnote{Here we ignore the \textit{multiple} measurement structure}, $p_S$ is the trusted labeled source distribution, and $p_T$ is the untrusted distribution which we \textit{treat} as fully unlabeled despite the presence of untrusted measurements. 

We evaluate ACE 0.1 and DivDis on two of the benchmark datasets, 
``Diamond in a Vault'' and ``Generated Stories'', 
using CodeGen-350M \citep{nijkamp2023codegenopenlargelanguage} and Pythia-1B \citep{biderman2023pythiasuiteanalyzinglarge} respectively. 
As before, we use validation loss to tune hyperparameters (loss weight and learning rate), and otherwise use the defaults established in the original benchmark. 
See Appendix~\ref{appendix:mtd-details} for a full list of hyperparameters.
Because the agent never tampers when the ground-truth concept $C_y(x)$ is true,  we only use $\mathbf{c} = (0,1)$ in the ACE loss. 
This asymmetric structure also distinguishes the target classifier from the measurement classifier, as the target classifier should predict false on instances where the hypotheses disagree. 
We compare ACE 0.1 and DivDis against the three measurement tampering detection methods.
See Appendix~\ref{appendix:mtd-details} for a description of the methods and training process 
(in particular on computing batch-dependent losses with gradient accumulation). 
Note the untrusted distribution has a fake positive rate of approximately 10\%.

\begin{figure}
    \centering
    \includegraphics[width=\columnwidth]{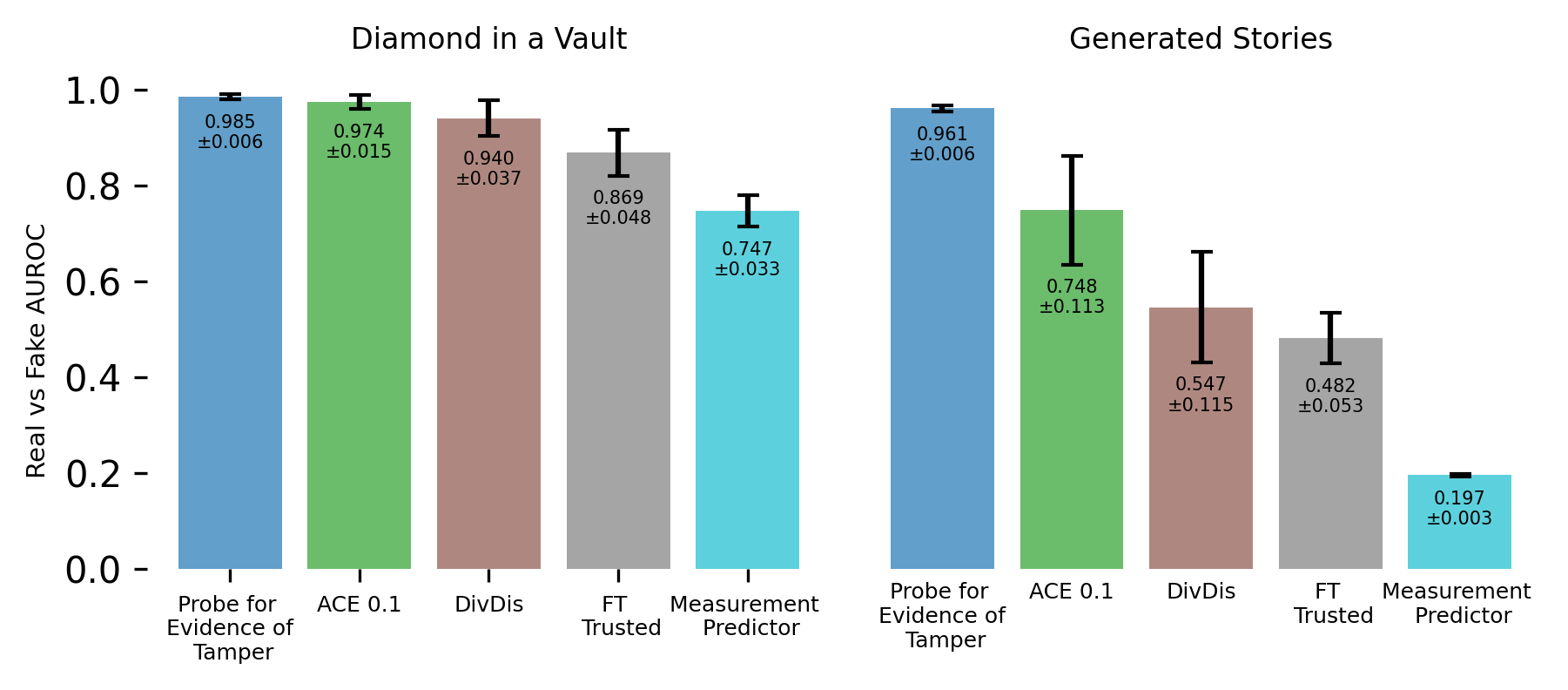}
    \caption{
        \textbf{Measurement Tampering Detection Real vs Fake AUROC}
        Mean and standard deviation of AUROC scores for detecting real versus fake positive outcomes across different methods, 
        averaged over 8 random seeds. ACE outperforms DivDis, and both outperform baseline methods, but Probing for Evidence of Tamper remains state-of-the-art.
    }
    \label{fig:mtd_results}
\end{figure}

Results are shown in Figure~\ref{fig:mtd_results}. 
We find that probing for evidence of tamper achieves the highest AUROC on both datasets. 
ACE achieves comparable performance on Diamond in a Vault (within 1\% AUROC), 
and substantially outperforms the baseline methods on Generated Stories. 
DivDis consistently outperforms the baseline methods but achieves lower performance than ACE, 
particularly on the Generated Stories dataset where the gap is most pronounced
\footnote{
    There is some divergence in our results and those reported in \citep{roger2023benchmarksdetectingmeasurementtampering}. 
    In particular, we find that Probing for Evidence of Tamper achieves higher AUROC (0.985 and 0.961) 
    for Diamond in a Vault and Generated Stories than previously reported (0.95 and 0.86). 
    We suspect this is due to subtle differences in the ``pretrained'' measurement predictor due to 
    random initialization and dataset ordering, though it is possible there is some more substantive difference. 
    In either case, we caution against direct comparisons between our results and those previously reported. 
}.

While neither ACE nor DivDis achieve state-of-the-art performance,
neither make use of the multiple measurement structure, 
suggesting concept extrapolation may be useful in cases where the agent
is sophisticated enough to never leave traces of failed tampering attempts. We also note that unlike prior experiments, measurement tampering induces an additional \textit{easy-to-hard} shift across the \textit{entire} untrusted distribution, potentially making consistent disagreement on instances where the ground-truth concepts agree more likely. Future work could address this concern by combining ACE with standard domain adaptation techniques for minimizing the distance between representations of domains (see \citep{ben2006analysis} and Appendix~\ref{appendix:related-work}). 

\section{Conclusion}
\label{section:conclusion}
We presented ACE, an approach for learning diverse generalizations of 
under-specified data by encouraging \textit{confident selective} disagreement. 
ACE achieves state-of-the-art performance on an array of complete spurious correlation benchmarks, and allows for natural, principled hyperparameter selection than prior disagreement-based ensemble methods. 
We also find ACE can perform well in measurement tampering detection without access to untrusted measurements,
demonstrating the potential applicability of ACE and concept-extrapolation more broadly to scalable oversight.

\subsection{Limitations and Future Work}
\label{section:future_work}
The present work contains two fundamental limitations, both of which we would be excited to see addressed in future work. 
First, ACE is sensitive to choice of mix rate lower bound. While we show 
validation loss degradation provides a useful signal for lower bound selection, our proposed 
thresholding method (20\% of the range in validation loss) is somewhat ad-hoc, and requires 
many trials to select the best value. 
Future work might address this limitation with out-of-distribution detection methods \citep{lang2023survey}
or use model uncertainty metrics to dynamically identify the mix rate \citep{rizve2021defensepseudolabelinguncertaintyawarepseudolabel}
(though such approaches would need to avoid flagging examples from different domains that are still correlation-aligned.)
Second, disagreement alone is not always sufficient to learn 
the intended generalization. In particular, hypotheses may learn to disagree 
on the basis of unrelated features as long as they can maintain strong performance 
on the labeled source distribution.
To address this limitation, future work might make use of techniques in domain adaptation (see related work in Appendix~\ref{appendix:related-work}), 
for example penalizing the distance between the source and target distributions \citep{long2015dan} or among instances of a given class \citep{Schroff_2015}. 

For future applications, we are especially excited about work building on our measurement 
tampering results, applying ACE more broadly in the context of weak-to-strong \citep{burns2023weak} and 
easy-to-hard \citep{hase2024unreasonableeffectivenesseasytraining} generalization, 
and deployment-time monitoring of untrusted models \citep{greenblatt2024aicontrolimprovingsafety, roger2023coup}.

Finally, we are excited about methods implementing confident and selective disagreement 
with different loss functions (e.g. soft-relaxations of ACE), 
and theoretical work analyzing expected error, convergence, and optimal solutions for ACE 
under various conditions.

\begin{ack}
This work was partially supported by the National Institute on Aging through grant P30AG073107 and partially funded by Aligned AI Limited.
\end{ack}

\bibliographystyle{plain}
\bibliography{references}

\newpage
\appendix

\section{Related Work}
\label{appendix:related-work}
\textbf{Spurious Correlations} occur when a non-essential attribute is 
correlated with the target attribute on the training distribution, and the inductive biases 
of the model favor the spurious attribute. Such inductive biases often favor 
``simpler'' attributes \citep{shah2020pitfallssimplicitybiasneural,valleperez2019deeplearninggeneralizesparameterfunction}, 
such as background \citep{ribeiro2016whyitrustyou} or texture \citep{geirhos2022imagenettrainedcnnsbiasedtexture}. As a result, 
spurious correlations are often referred to as ``shortcut learning'' \citep{Geirhos_2020}. 
Spurious correlations create \textit{biased} models, which can suffer performance degradation 
\textit{in-distribution} (when the combined ``availability and predictivity'' of the spurious attribute
are higher than that of the target attribute \citep{hermann2024on}), 
under \textit{subpopulation shift} (when a minority group of bias-conflicting examples becomes more prominent in the test distribution \citep{sagawa2020distributionallyrobustneuralnetworks}), 
or under \textit{domain shift} (when the new domain samples from a previously unseen distribution of bias-conflicting instances \citep{krueger2021outofdistributiongeneralizationriskextrapolation}).







\textbf{Subpopulation Shift} is a type of distribution shift where small subpopulation (a ``minority group'') 
present on the training distribution becomes more frequent on the test distribution (these minority groups 
often consist of correlation-conflicting examples). Some methods for handling subpopulation shift rely on 
labeled access to subpopulation groups
\citep{sagawa2020distributionallyrobustneuralnetworks, kirichenko2023layerretrainingsufficientrobustness,nam2022spreadspuriousattributeimproving,idrissi2022simpledatabalancingachieves,yao2022improvingoutofdistributionrobustnessselective}, 
while other approaches attempt to leverage the inherent \textit{difficulty} of the minority group examples
to automatically identify and up-weight them 
\citep{nam2020learningfailuretrainingdebiased,liu2021justtraintwiceimproving,lee2021learningdebiasedrepresentationdisentangled,creager2021environmentinferenceinvariantlearning,duchi2022distributionallyrobustlosseslatent}. 
Rather than inducing the subpopulation shift, methods for handling subpopulation shift are typically bench-marked by ``worst-group'' performance. 
Note that while subpopulation shift often involved spurious correlations, other types of shifts are 
also covered within the framework (including class and attribute imbalances \citep{yang2023changehardcloserlook}). 
For class imbalance, traditional methods like reweighting and up-sampling can be effective \citep{Japkowicz2000TheCI}, 
though more recent work focused on using label distribution aware losses to increase classification margins 
on minority groups \citep{cao2019learningimbalanceddatasetslabeldistributionaware,cui2019classbalancedlossbasedeffective,yang2020rethinkingvaluelabelsimproving}.

\textbf{Domain Generalization} is the problem of generalizing to unseen domains using training data from multiple domains. 
Approaches typically try to learn \textit{invariant} representations across domains, 
such that the learned representations and predictor remain stable in novel domains 
\citep{li2018domaingeneralization,arjovsky2020invariantriskminimization,krueger2021outofdistributiongeneralizationriskextrapolation,chevalley2022invariantcausalmechanismsdistribution,yao2022improvingoutofdistributionrobustnessselective}. 
While most approaches rely on domain labels, \citep{creager2021environmentinferenceinvariantlearning} 
attempts to infer domain partitions. Similarly to subpopulation shift, domain generalization 
often, but not always, involves spurious correlations. The distinction between 
sup-population shift is not always clear (e.g. some works on domain generalization evaluate 
methods on subpopulation shift benchmarks \citep{krueger2021outofdistributiongeneralizationriskextrapolation,creager2021environmentinferenceinvariantlearning}).
We consider generalizing from completely spuriously correlated data to be a particular 
instance of domain generalization - and methods that use ``internal'' diversity metrics 
to learn diverse or de-biased ensembles without target domain data
\citep{teney2022evadingsimplicitybiastraining,yang2022chromavaemitigatingshortcutlearning,tiwari2023overcomingsimplicitybias} 

% ...

\textbf{Domain Adaptation} is a relaxed version of domain generalization which assumes 
\textit{unlabeled} access to the target domain. Early foundation work established 
formal bounds on the target error as a function of the source error and divergence 
between the source and target distributions \citep{ben2006analysis,mansour2009domain,BenDavid2010ATO}. 
This theoretical work motivated the development of methods that explicitly penalize  
the distance between the source and target distributions, using maximum mean discrepancy 
\citep{gretton2012kernel,long2015dan,long2016jan} or correlation distance \citep{sun2015return, sun2016deepcoralcorrelationalignment}. 
Similarly motivated work proposes minimizing source-target discrepancy using adversarial training 
\citep{ganin2016domain,tzeng2017adversarialdiscriminativedomainadaptation,hoffman2017cycadacycleconsistentadversarialdomain,shen2018wasserstein}, 
learning representations that maximally confuse a discriminator trained to distinguish between 
source and target domains. 
While not central examples of domain adaptation, we consider ACE and other approaches 
that use unlabeled target data to learn diverse ensembles 
(DivDis \citep{lee2023diversifydisambiguatelearningunderspecified} and D-BAT \citep{pagliardini2022dbat}),
to be a class of domain adaptation methods. 

\textbf{Diverse Ensembles} were recognized as desirable for robust generalization early 
in the development of modern machine learning \citep{krogh1994neural}. 
Modern deep ensembles typically use different initializations and dataset orderings to 
produce diverse predictions \citep{lakshminarayanan2017simplescalablepredictiveuncertainty}. 
However, ensembles often converge to similar local optima when the inductive biases of the model
strongly favor a given solution (e.g. using a particular spurious attribute). To promote 
more diverse solutions, methods explicitly promote diversity by following different 
eigenvectors of the hessian \citep{parkerholder2020ridgeriderfindingdiverse}, or adversarially 
learning diverse latent representations 
\citep{sinha2020diversityinducinginformationbottleneck, rame2021dicediversitydeepensembles}. 
Most similar to ACE, DivDis and D-BAT explicitly penalize mutual information or agreement 
in the predictions of the ensemble members 
\citep{lee2023diversifydisambiguatelearningunderspecified,pagliardini2022dbat}, 
allowing them to handle extreme (in some cases complete) spurious correlations. 
Diverse ensembles have also been used to improve scalable oversight \citep{amodei2016concreteproblemsaisafety}, 
acting as an uncertainty metric for active learning in reinforcement learning from human feedback \citep{christiano2017deep} 
and mitigating over-optimization against reward models \citep{coste2024rewardmodelensembleshelp}. 

\textbf{Self-Training} is an approach to semi-supervised learning which uses (hardened) model 
predictions as training targets. Popular methods like pseudo-labeling \citep{Lee2013PseudoLabelT} 
and entropy regularization \citep{grandvalet2004semi} rely on the \textit{cluster assumption} 
\citep{chapelle2005lowdensity} - that data form clusters with each elements of each cluster sharing 
the same class and classes separated by \textit{low density} regions. 
Self-training methods have achieved state-of-the art performance \citep{xie2020self}, 
but pure pseudo-labeling approaches can cause instability and reinforce inaccurate predictions. 
To overcome these issues, prior work suggests confidence \citep{CascanteBonilla2020CurriculumLR} and 
uncertainty \citep{rizve2021defensepseudolabelinguncertaintyawarepseudolabel} based thresholding, 
as well as a gradual schedule for the pseudo-label loss \citep{Lee2013PseudoLabelT}. 
Some work has successfully applied self-training to domain-adaption 
\citep{french2018selfensemblingvisualdomainadaptation,shu2018dirt,kumar2020understanding}.
Most relevant to our work, \citep{chen2020self} proves that under certain conditions 
self-training can improve robustness to spurious correlation (though they assume 
the default classifier favors the ground-truth feature).

\section{Proper Scoring Under Different Mix Rates}
\label{appendix:mix-rate-proofs}
A scoring rule $S(p,q)$ evaluates how well a distribution $p$ approximates $q$. A scoring rule is \textit{proper} if the true distribution $q$ achieves a global minimum: $S(q, q) \leq S(p, q)$. In the context of diverse generalization, our true distribution $q$ is a joint distribution identical to the set of $F$ concepts: 
\begin{align}
q(\mathbf{y}|x) = 
\begin{cases}
1 & \text{if}\ \mathbf{y}= \mathbf{C}(x) \\ 
0 & \text{otherwise}
\end{cases}
\end{align}
with $\mathbf{y} = (y_1, \dots, y_n), \mathbf{C}(x) = (C_1(x), \dots, C_F(x))$. Note that our ensemble $\boldsymbol{\theta}$ can similarly be expressed as a distribution over concepts, with $p_{\boldsymbol{\theta}}(\mathbf{y}|x) = (p_{\theta_1}(y_1|x),  \dots,  p_{\theta_F}(y_F|x))$. 

 Note that (under complete search) proper scoring is \textit{necessary} but not \textit{sufficient} for finding the true distribution. In particular, without strictly proper scoring ($S(q, q) < S(p,q)$ for $ p \neq q$), we may find global minima arbitrarily far from the true distribution. For example, vanilla ERM is (trivially) a proper scoring rule for diverse generalization, but insufficient for finding good approximations to $q$. While stochastic gradient descent is not complete, in practice we expect losses that are proper scoring rules to be more effective. 

In the following subsections, we show that ACE is a proper scoring rule under all mix rates above the assumed lower bound. In contrast, we show DivDis and D-BAT are only proper under a unique mix rate. For simplicity, we only cover the case of $F=2$. 

For ease of notation, we let $p_{c_i}^y = p(C_i(x)=y)$ denote the marginal probability that the $i-th$ concept equals $y$,  $p_{c_i}$ denote the full marginal distribution, and $p_{c_1 c_2}$ denote the joint distribution over both concepts. 
\subsection{DivDis}
\label{appendix:divdis_mix_rate_proof}

\begin{proposition}
DivDis is a proper scoring rule if and only if the mix rate $r$ on $q$ equals the specific value $r^* = q_{c_1}^0 q_{c_2}^1 + q_{c_1}^1 q_{c_2}^0$
\end{proposition}

\begin{proof}
Recall that DivDis penalizes mutual information between concept distributions: 
\begin{equation}
\mathcal{L}_{\text{MI}}(p_{c_1}, p_{c_2}) = D_{\text{KL}}(p_{c_1, c_2} || p_{c_1} \times p_{c_2})
\end{equation}
and note $D_{KL}(q,p) \geq 0$. If DivDis is a proper scoring rule, then by definition $q$ must attain this lower bound, i.e.  $\mathcal{L}_{\text{MI}}(q_{c_1}, q_{c_2}) = 0$. The other direction also trivially holds. Thus concepts $c_1$ and $c_2$ have $0$ mutual information under $q$ if and only if $r^* = q_{c_1}^0 q_{c_2}^1 + q_{c_1}^1 q_{c_2}^0$. 
First, recall that if two variables have 0 mutual information, they must be statistically independent, i.e. 
\begin{align*}
q_{c_1, c_2}^{0,1} &= q_{c_1}^0 q_{c_2}^1 \\
q_{c_1, c_2}^{1,0} &= q_{c_1}^1 q_{c_2}^0 \\
\end{align*}
Recall that the mix rate is given by $r = \mathbb{E}_q[\mathbf{1}_{C_1(x) \neq C_2(x)}] = q_{c_1, c_2}^{0,1} + q_{c_1, c_2}^{1,0}$. 
Substituting in the above, we get 
\begin{align*}
r &= q_{c_1, c_2}^{0,1} + q_{c_1, c_2}^{1,0} \\ 
    &= q_{c_1}^0 q_{c_2}^1 + q_{c_1}^1 q_{c_2}^0 = r^*
\end{align*}
Showing the other direction, assume that $r = r^* = q_{c_1}^0 q_{c_2}^1 + q_{c_1}^1 q_{c_2}^0$. Then by the definition of mix rate and the marginal covariance decomposition, we must have zero covariance between $c_1$ and $c_2$. Thus the concepts are statistically independent, with $q_{c_1, c_2} = q_{c_1} p_{c_2}$, implying $\mathcal{L}_{MI}(q_{c_1}, q_{c_2}) = 0$ 
\end{proof}

Note that for evenly balanced marginal concept distributions $p_{c_i}^0 = p_{c_i}^1 = 0.5$, 
$r^* = 0.5$.
See Appendix \ref{appendix:dataset_class_distributions} for computations 
of the ``optimal'' mix rate $r^*$ given the marginal concept distributions 
for all settings of the realistic datasets used in this work.

\subsection{D-BAT}

\begin{proposition}
D-BAT is a proper scoring rule if and only if the mix rate $r=1$
\end{proposition}

\begin{proof}
First, recall that D-BAT penalizes the expected negative log-probability of disagreement between concept distributions on each instance:
\begin{equation*}
\mathcal{L}_{\mathcal{A}}(p_1, p_2) =  \mathbb{E}\left[-\log \left(p_{c_1|x}^0 p_{c_2 |x}^1 + p_{c_1|x}^1 p_{c_2|x}^0\right)\right]
\end{equation*}

Since the domain of the log argument is a probability and $-\log(x)$ is strictly decreasing, this loss is minimized when the probability of disagreement is 1 for all instances, with $\mathcal{L}_{\mathcal{A}}(p_1, p_2) \geq 0$. Thus if $\mathcal{L}_{\mathcal{A}}$ is a proper scoring rule, then $q_{c_1|x}^0 q_{c_2 |x}^1 + q_{c_1|x}^1 q_{c_2|x}^0 =1$ for all $x$. But if this equality holds, then $1=\mathbb{E}[\mathbf{1}_{C_1(x) \neq C_2(x)}]=r$. Going the other way, if $r=1$ then the probability of disagreement is 1 on all inputs, so the expectation of the negative log is 0. 
\end{proof}

\subsection{ACE}

\begin{proposition}
ACE-TopK is a proper scoring rule if and only if each group mix rate lower bound does not exceed the true mix rate, i.e. $\forall g' \in G': \bar{r}_{g'} \leq r_{g'}$
\end{proposition}

\begin{proof}
To avoid complications with batch sizing, we assume the inputs $x$ are sampled uniformly from a discrete set $\mathcal{D}$ of size $N$. A simplified version of the ACE loss is given below:
\begin{align*}
\mathcal{L}_{\text{ACE}}(p) &=  \mathbb{E}\left[m_{i}^{(0,1)} \cdot -( \log{p_{c_1|x}^0} + \log{p_{c_2|x}^1}) + m_{i}^{(1,0)} \cdot -( \log{p_{c_1|x}^1} + \log{p_{c_2|x}^0})\right] \\ 
 m_{i}^\mathbf{y} &\leftarrow 
    \begin{cases}
        1 & \mbox{ ... if } i\in \mbox{topk}(k_{\mathbf{y}},\mathcal{P}^\mathbf{y}) \\
        0 & \mbox{ ... otherwise}
    \end{cases} \\
\end{align*}
with $k_{\mathbf{y}} = \lfloor N \cdot \bar{r}_{\mathbf{y}} \rfloor$ and $\mathcal{P}^{\mathbf{y}}$ the set of probabilities each instance has the concept labels $\mathbf{y}$. Note the slight abuse of notation, replacing the group index $g$ with the corresponding concept values $\mathbf{y}^g$ for clarity. This loss is bounded below by $0$, with the bound obtained when the probabilities are 1 for all unmasked instances.

First assume the true mix rate for each group is greater than the lower bound. Then $k_{\mathbf{y}} = \lfloor N \cdot \bar{r}_{\mathbf{y}} \rfloor  \leq |\{x| \mathbf{c}(x_j)=\mathbf{y}\}|$. Because the true distribution $q$ is able to perfectly sort instances, the top-k most likely instances to have concept labels $\mathbf{y}$ will actually have such labels, thus achieving 0 loss. 

Going the other way, assume $\mathcal{L}_{\text{ACE}}$ is a proper scoring rule and that the lower bound does not hold, i.e.  $k_{\mathbf{y}} > |\{x| \mathbf{c}(x_j)=\mathbf{y}\}|$ for disagreeing labels $\mathbf{y}$. Then the top-k function \textit{must} select an instance $x$ such that $\mathbf{c}(x) \neq \mathbf{y}$ which will accrue non-zero loss under $q$, contradicting our initial assumption. 
\end{proof}

\section{Complete Correlation Datasets}
\label{appendix:cc-datasets}

\textbf{Gaussian Blob 2D Grid}: A modified version of the toy 2D grid dataset 
introduced by \citep{lee2023diversifydisambiguatelearningunderspecified}, 
where instead of uniformly distributing data across quadrants, 
we sample data points from 2D gaussian distributions centered in the middle of each quadrant, 
with a standard deviation of 0.01, more closely resembling binary features 
separated by low density regions.

\textbf{FashionMNIST-MNIST}: A dataset of concatenated FashionMNIST 
\citep{xiao2017fashionmnistnovelimagedataset} and MNIST \citep{deng2012mnist} images 
introduced by \citep{pagliardini2022dbat}, 
following the ``dominos'' format introduced by \citep{shah2020pitfallssimplicitybiasneural} 
(see CIFAR-MNIST below). 
The source distribution contains MNIST $0$'s concatenated with FashionMNIST tops, 
and MNIST $1$'s concatenated with FashionMNIST bottoms. 
The target dataset contains some instances of MNIST $0$'s concatenated to FashionMNIST bottoms, 
and MNIST $1$'s concatenated to FashionMNIST tops. 

\textbf{CIFAR-MNIST}: A dataset of concatenated CIFAR-10 and MNIST images 
introduced by \citep{shah2020pitfallssimplicitybiasneural}. 
The source distribution contains MNIST $0$'s concatenated with CIFAR cars, 
and MNIST $1$'s concatenated with CIFAR trucks. 
The target dataset contains some instances of MNIST $0$'s concatenated to CIFAR trucks, 
and MNIST $1$'s concatenated to CIFAR cars. 

\textbf{Waterbirds-CC}: A modified version of Waterbirds \citep{sagawa2020distributionallyrobustneuralnetworks} 
introduced by \citep{lee2023diversifydisambiguatelearningunderspecified} 
to evaluate complete spurious correlation. Waterbirds-CC contains only images of land birds on land 
and waterbirds on water in the source distribution, 
and some instances of land birds on water and waterbirds on land in the target distribution. 

\textbf{CelebA-CC}: A modified version of the CelebA dataset \citep{liu2015faceattributes}, 
CelebA-CC contains only images of blond woman and dark-haired men on the source distribution, 
but some instances of blond men and dark-haired woman on the target distribution 
(where the ground truth feature is hair color and spurious feature is gender). 

\textbf{Multi-NLI-CC-Binary}: A modified version of Multi-NLI \citep{Williams2018} 
Similar to Multi-NLI-CC introduced by \citep{lee2023diversifydisambiguatelearningunderspecified}, 
but groups entailment and neutral into a single label and filters out instances of 
contradictions without negation from the source distribution. Thus for the source distribution, 
all instances contain either entailment or neutral with no negation words, or 
contradictions with negation words. The target distribution contains all four groups.


\section{Dataset Class Distributions}
\label{appendix:dataset_class_distributions}
As described in Section~\ref{sec:mix_rate_robustness}, the three ``realistic'' datasets 
used to evaluate diverse generalization methods (Waterbirds-CC, CelebA-CC, Multi-NLI-CC-Binary) 
have class and group imbalances, which can impact the performance of methods and 
our interpretation of the results. Below we provide the


\begin{table}[h]
    \centering
    \begin{tabular}{lcc}
    \toprule
    Dataset & $p^{(0,1)}$ & $p^{(1,0)}$ \\
    \midrule
    Waterbirds-CC & 0.4 & 0.1 \\
    CelebA-CC & 0.49 & 0.01 \\
    Multi-NLI-CC-Binary & 0.03 & 0.47 \\
    \bottomrule
    \end{tabular}
    \vspace{0.5em}
    \caption{Disagreement group distributions across datasets}
    \label{tab:minority_group_dist}
\end{table}

\begin{table}[h]
    \centering
    \begin{tabular}{lccc}
    \toprule
    Dataset & Mix Rate & $(p_{c_1}^0, p_{c_2}^0)$ & $p_{c_1}^0 (1-p_{c_2}^0) + (1-p_{c_1}^0) p_{c_2}^0$ \\
    \midrule
    Waterbirds-CC & - & (0.79, 0.50) & 0.50 \\
    \midrule
    & 0.10 & (0.76, 0.69) & 0.40 \\
    & 0.25 & (0.80, 0.56) & 0.46 \\
    CelebA-CC & 0.50 & (0.86, 0.38) & 0.59 \\
    & 0.75 & (0.92, 0.20) & 0.75 \\
    & 1.00 & (0.98, 0.02) & 0.96 \\
    \midrule
    & 0.10 & (0.84, 0.93) & 0.21 \\
    & 0.25 & (0.71, 0.93) & 0.32 \\
    Multi-NLI-CC-Binary & 0.50 & (0.50, 0.93) & 0.50 \\
    & 0.75 & (0.28, 0.93) & 0.69 \\
    & 1.00 & (0.06, 0.94) & 0.88 \\
    \bottomrule
    \end{tabular}
    \vspace{0.5em}
    \caption{Marginal class distributions and zero mutual information mix rates across datasets and mix rate conditions}
    \label{tab:marginal_dist}
\end{table}

\section{Experiment Details}
\label{appendix:experiment_details}
\subsection{Architecture and Training Details}
In all experiments, we train 2 classification heads. 
On image datasets, we use a shared ResNet50 \citep{he2015deepresiduallearningimage} backbone, 
for text, we use a shared BERT \citep{devlin2019bertpretrainingdeepbidirectional} backbone.
For the Toy Grid task, we use a 2 hidden layer multi-layer perception with a hidden dimension of 40, and 
only select from models that achieve 100\% source accuracy.
Because D-BAT requires training one model at a time while freezing the others, 
we use separate backbones and half the batch size for each model, 
training each model for half the total allotted epochs.  
For ACE, the mix rate lower bound is incremented 
linearly (starting from 0) every epoch. 

We train all models on  NVIDIA A6000's (48GB), though the image models can be trained with fewer resources.

\subsection{Hyperparameters}
Learning rate, optimizer, weight decay, and auxiliary loss weight are all tuned on each 
dataset, though we only notice substantial variation in the auxiliary loss weight. Tuning 
auxiliary loss weight according to total validation loss (with an auxiliary loss weight of 1) 
may seem counterintuitive, but is motivated by work on label-distribution aware losses 
\citep{cao2019learningimbalanceddatasetslabeldistributionaware, cui2019classbalancedlossbasedeffective}.
Initial experiments found little variation in optimal parameters across mix rates, 
so auxiliary loss is tuned on default dataset mix rates (typically close to 0.5).

Below we describe the full list of hyperparameters:

\subsubsection{Robustness to Mix Rate Variations Experiments}

Below are the hyperparameters used for the robustness to mix rate variations experiments.
All parameters are tuned according to total validation loss (without using any target labels).

For text datasets, we also use a half-cosine learning rate schedule with 
a linear warmup period of 0.05 of the total training steps.

\begin{table}[h]
    \centering
    \begin{tabular}{lcccccc}
    \toprule
    Dataset & Learning Rate & Optimizer & Weight Decay & Epochs & $N_S$ & $N_T$ \\
    \midrule
    Vision & 1e-4 & AdamW & 1e-3 & 5 & 32 & 64 \\ 
    Text & 1e-5 & AdamW & 1e-3 & 2 & 32 & 64 \\
    Toy Grid & 1e-3 & AdamW/SGD* & 1e-2 & 100 & 32 & 128 \\
    \bottomrule
    \end{tabular}
    \vspace{0.5em}
    \caption{Optimization hyperparameters for different datasets. \\
        \small *only TopK uses SGD on Toy Grid
    }
    \label{tab:opt_params}
\end{table}
    
\begin{table}[h]
    \centering
    {\small
    \begin{tabular}{lcccccc}
    \toprule
    Method & ToyGrid & FMNIST-MNIST & CIFAR-MNIST & Waterbirds-CC & CelebA-CC & Multi-NLI-CC-Binary \\
    \midrule
    TopK 0.1 & 1.5 & 1.0 & 1.5 & 8.0 & 2.5 & 6.0 \\
    TopK 0.5 & 1.0 & 1.0 & 1.0 & 3.0 & 1.5 & 6.0 \\
    DivDis & 1.0 & 1.5 & 1.5 & 6.0 & 1.5 & 64.0 \\
    D-BAT & 0.3 & 0.01 & 0.3 & 0.15 & 0.3 & 0.1 \\
    \bottomrule
    \end{tabular}
    }
    \vspace{0.5em}
    \caption{Auxiliary loss weights for different methods and datasets}
    \label{tab:aux_weights}
\end{table}

\subsubsection{Measurement Tampering Experiments}
Both DivDis and TopK 0.1 use a learning rate of 2e-5 for Diamonds and 
3e-6 for Generated Stories. Below is the table of auxiliary loss 
weights for each: 

\begin{table}[h]
    \centering
    \begin{tabular}{lcc}
    \toprule
    Method & Diamonds & Generated Stories \\
    \midrule
    TopK 0.1 & 1.0 & 2.5 \\
    DivDis & 3.0 & 3.0 \\
    \bottomrule
    \end{tabular}
    \vspace{0.5em}
    \caption{Auxiliary loss weights for measurement tampering experiments}
    \label{tab:tampering_weights}
\end{table}

\section{Robustness to Mix Rate Variations Futher Results}
\label{appendix:mix-rate-results}

\subsection{Disaggregated Results}
\begin{figure}[h]
    \centering
    \begin{subfigure}[b]{0.75\textwidth}
        \centering
        \includegraphics[width=\textwidth]{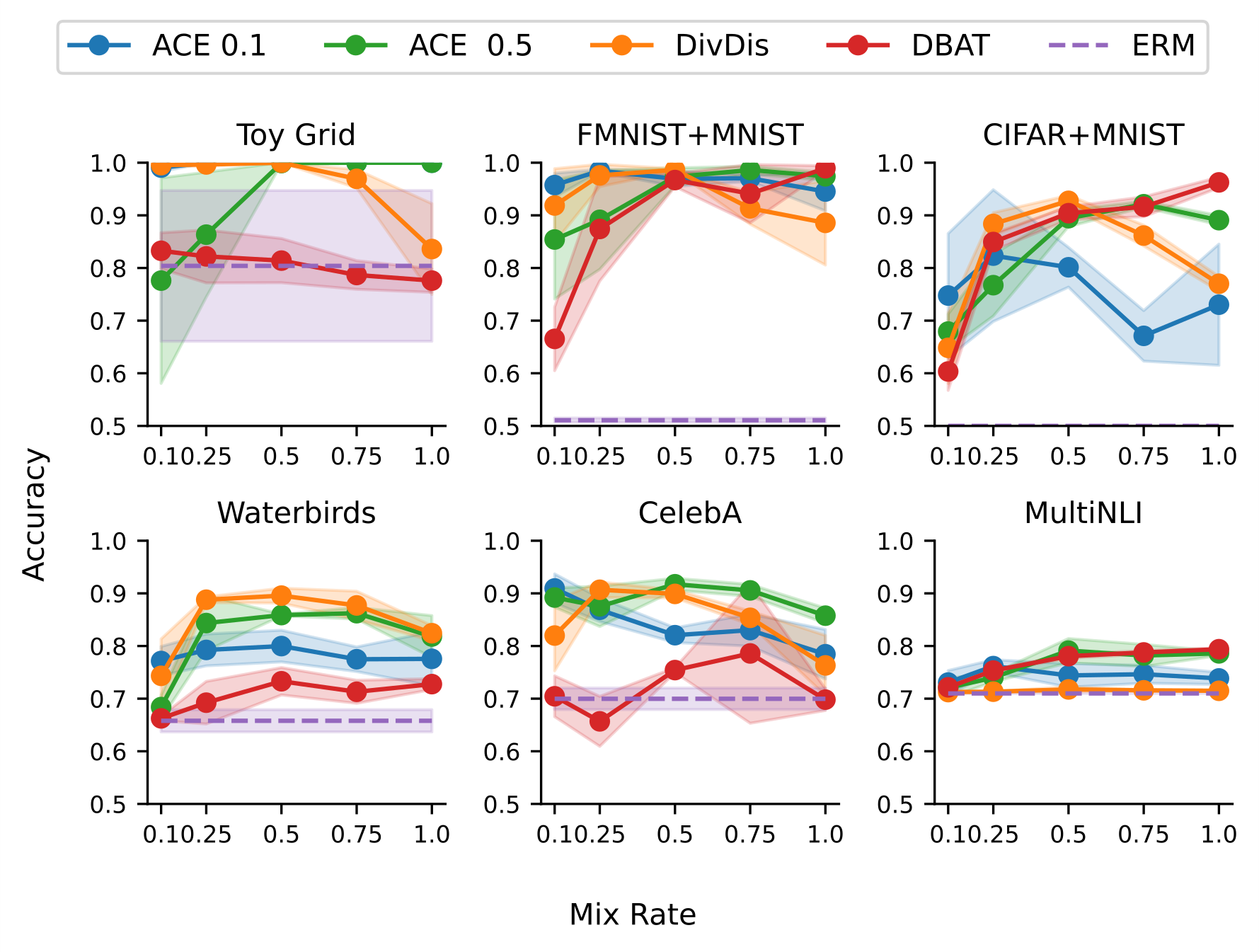}
        \label{fig:cc_mix_rate_acc}
    \end{subfigure}

    \begin{subfigure}[b]{0.75\textwidth}
        \centering
        \includegraphics[width=\textwidth]{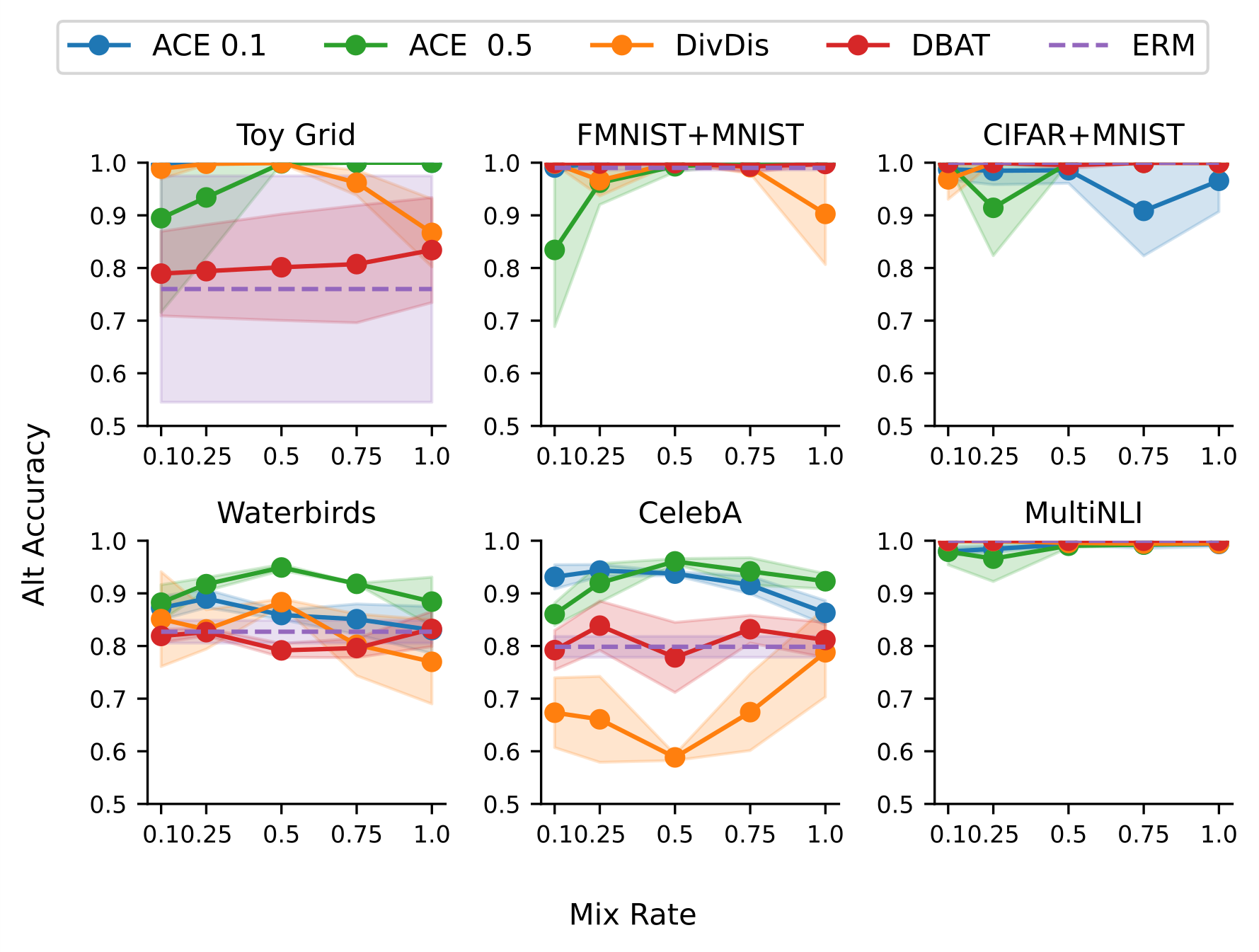}
        \label{fig:cc_mix_rate_alt}
    \end{subfigure}
    \caption{
        \textbf{Accuracy Metrics Across Mix Rates}
        Mean (line) and standard deviation (shaded area) of group truth 
        and spurious accuracy metrics across mix rates (taken over 3 seeds). 
        Results largely align with the main results. Most notably, 
        DivDis achieves higher ground-truth accuracy than spurious accuracy 
        on Waterbirds and CelebA, even though both generalizations
        are equally valid and the spurious features are prefered by ERM.
    }
    \label{fig:cc_mix_rate_combined}
\end{figure}

\begin{figure}[ht]
    \centering
    \includegraphics[width=0.75\textwidth]{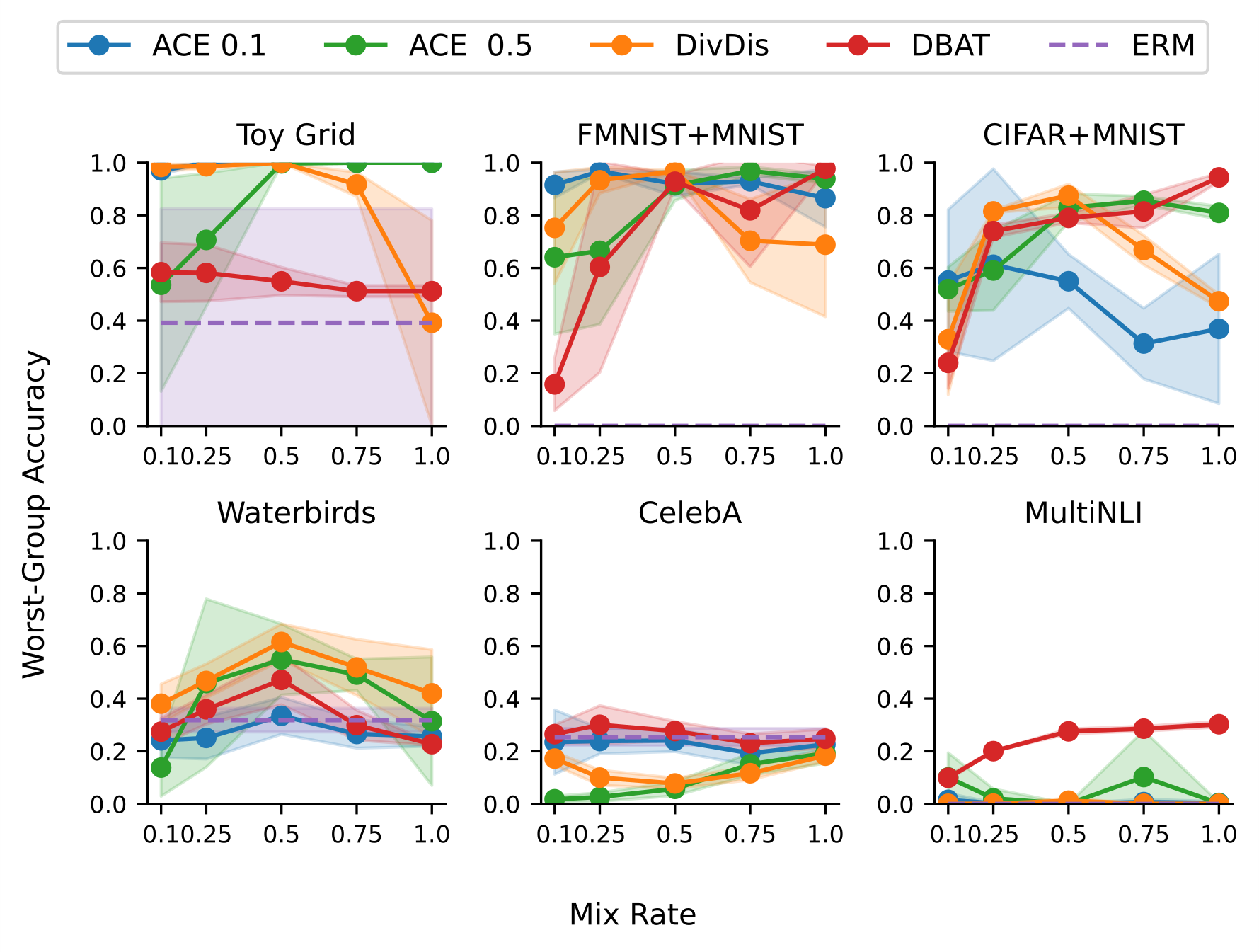}
    \caption{
        \textbf{Worst-Group Accuracy Across Mix Rates}
        Mean (line) and standard deviation (shaded area) of worst-group 
        accuracy across mix rates (taken over 3 seeds). Again results largely 
        conform to the main body findings, though TopK 0.5 approximately 
        pareto dominates all other methods on the realistic datasets (Waterbirds, 
        CelebA, and MultiNLI).
    }
    \label{fig:cc_mix_rate_worst}
\end{figure}

In figures \ref{fig:cc_mix_rate_combined}, and \ref{fig:cc_mix_rate_worst}, 
we present disaggregated results for the robustness to mix rate variations, 
showing average ground truth and spurious accuracy, as well as worst-group
ground truth accuracy. 

Note that class imbalances in the realistic datasets also imply that the ``0 mutual information'' 
point is not always at a mix rate of 0.5, potentially effecting the performance of 
DivDis. See Appendix~\ref{appendix:dataset_class_distributions} for the the class distributions
and ``0 mutual information'' points for each dataset.



\subsection{Mix Rate Scheduling Ablation}
\begin{figure}[ht]
    \centering
    \includegraphics[width=0.75\textwidth]{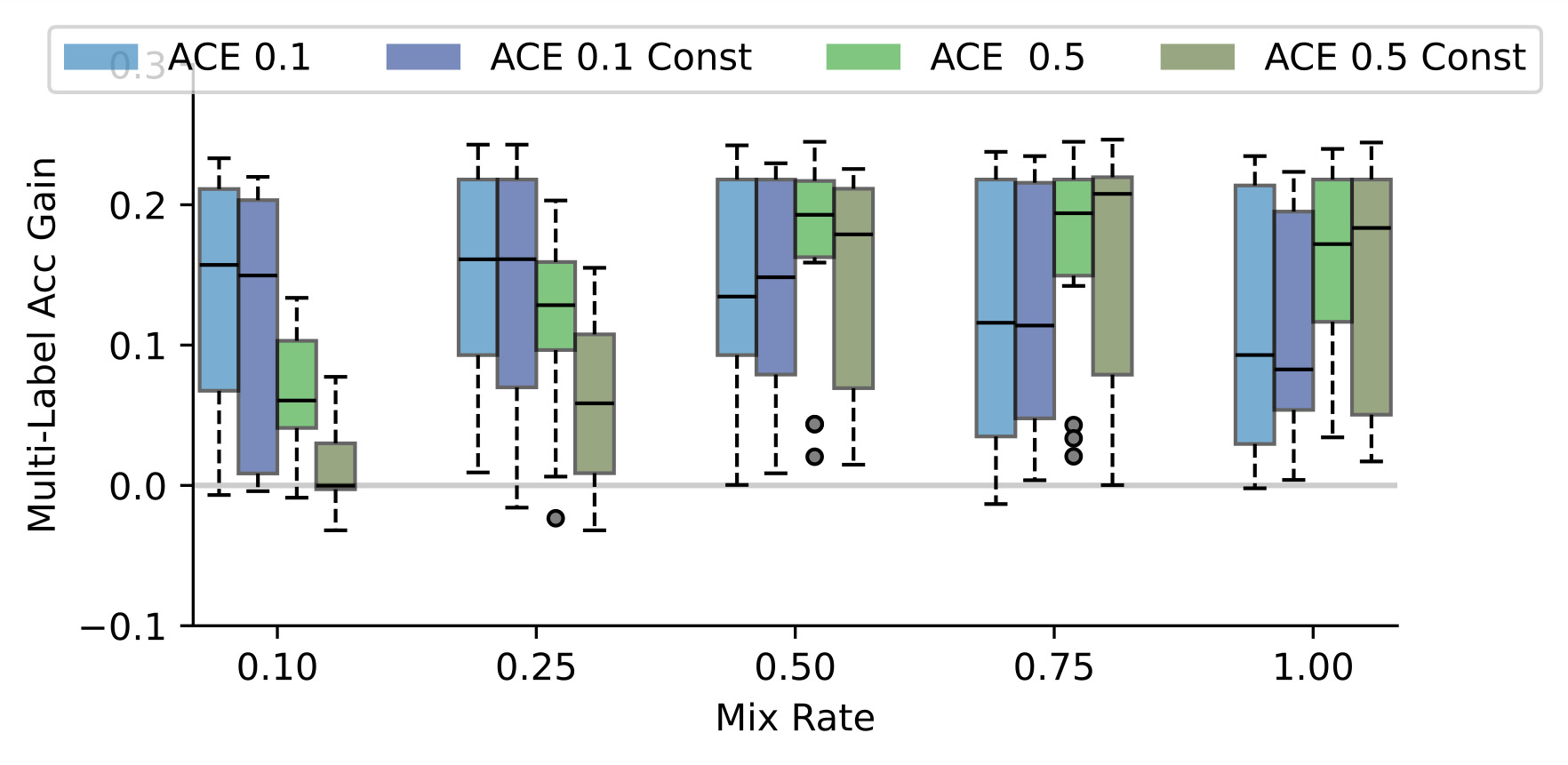}
    \caption{
        \textbf{(Scheduling Ablation) Aggregated Multi-Label Accuracy Across Mix Rates}
        Box plots of multi-label accuracy gain over ERM, 
        averaged over 6 datasets and 3 random seeds, 
        for TopK 0.1 and 0.5 with and without mix rate scheduling
        Mix rate scheduling substantially improves the performance of TopK 0.5 on low mix rates, 
        but provides only a modest improvement to TopK 0.1. 
    }
    \label{fig:cc_mix_rate_avg_ablat_acc_ablation}
\end{figure}

\begin{figure}[ht]
    \centering
    \begin{subfigure}[b]{0.7\textwidth}
        \centering
        \includegraphics[width=\textwidth]{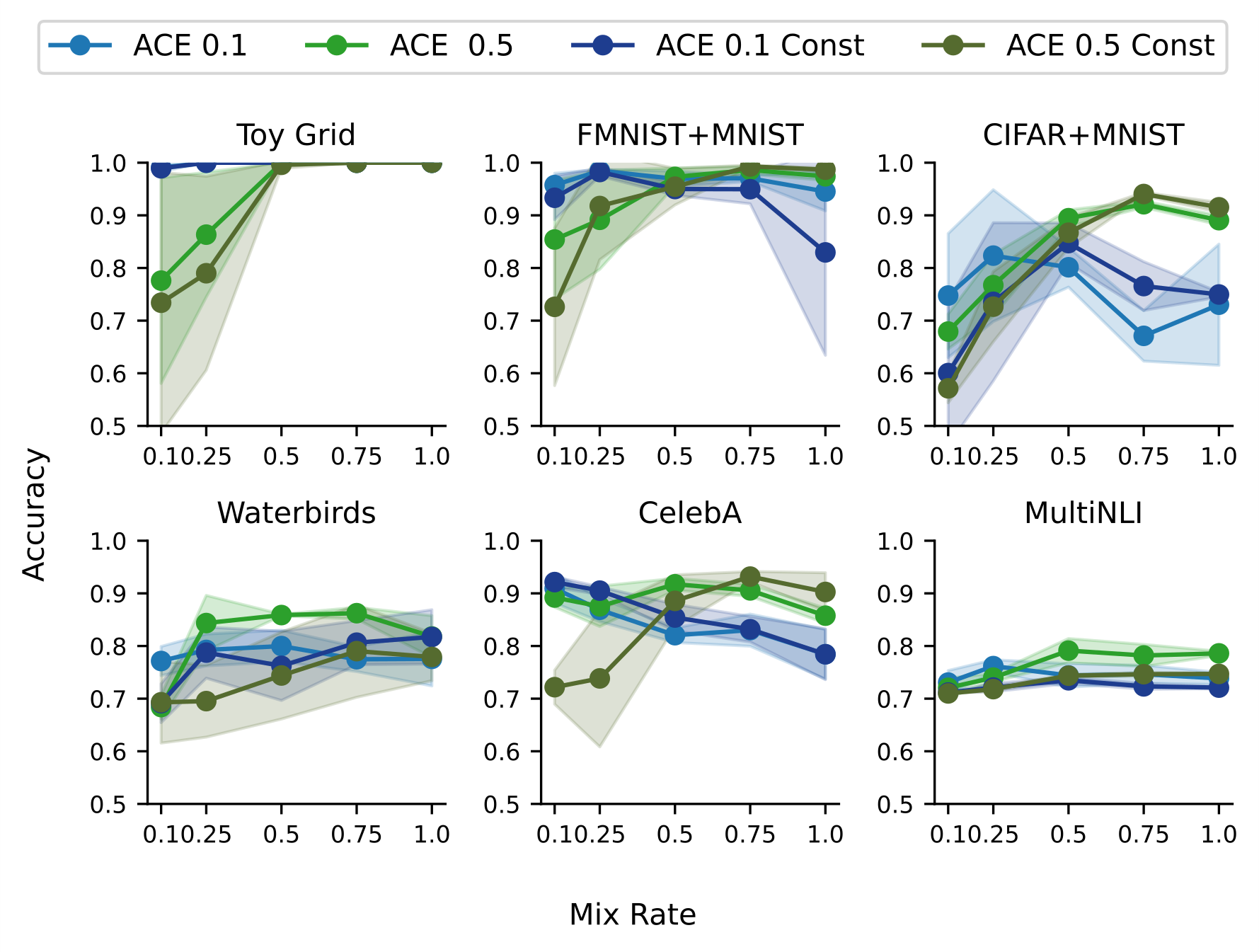}
    \end{subfigure}

    \begin{subfigure}[b]{0.7\textwidth}
        \centering
        \includegraphics[width=\textwidth]{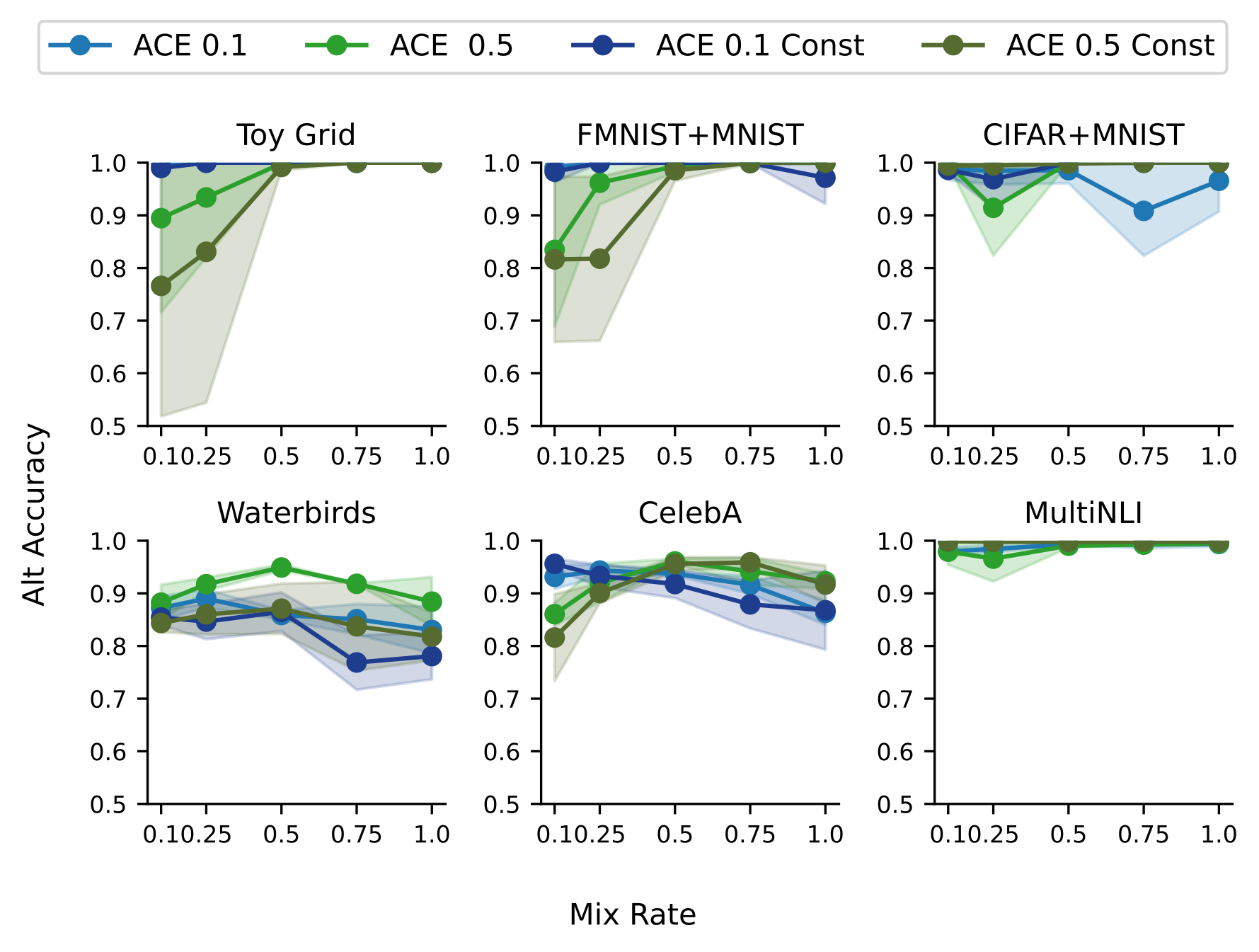}
    \end{subfigure}

    \begin{subfigure}[b]{0.7\textwidth}
        \centering
        \includegraphics[width=\textwidth]{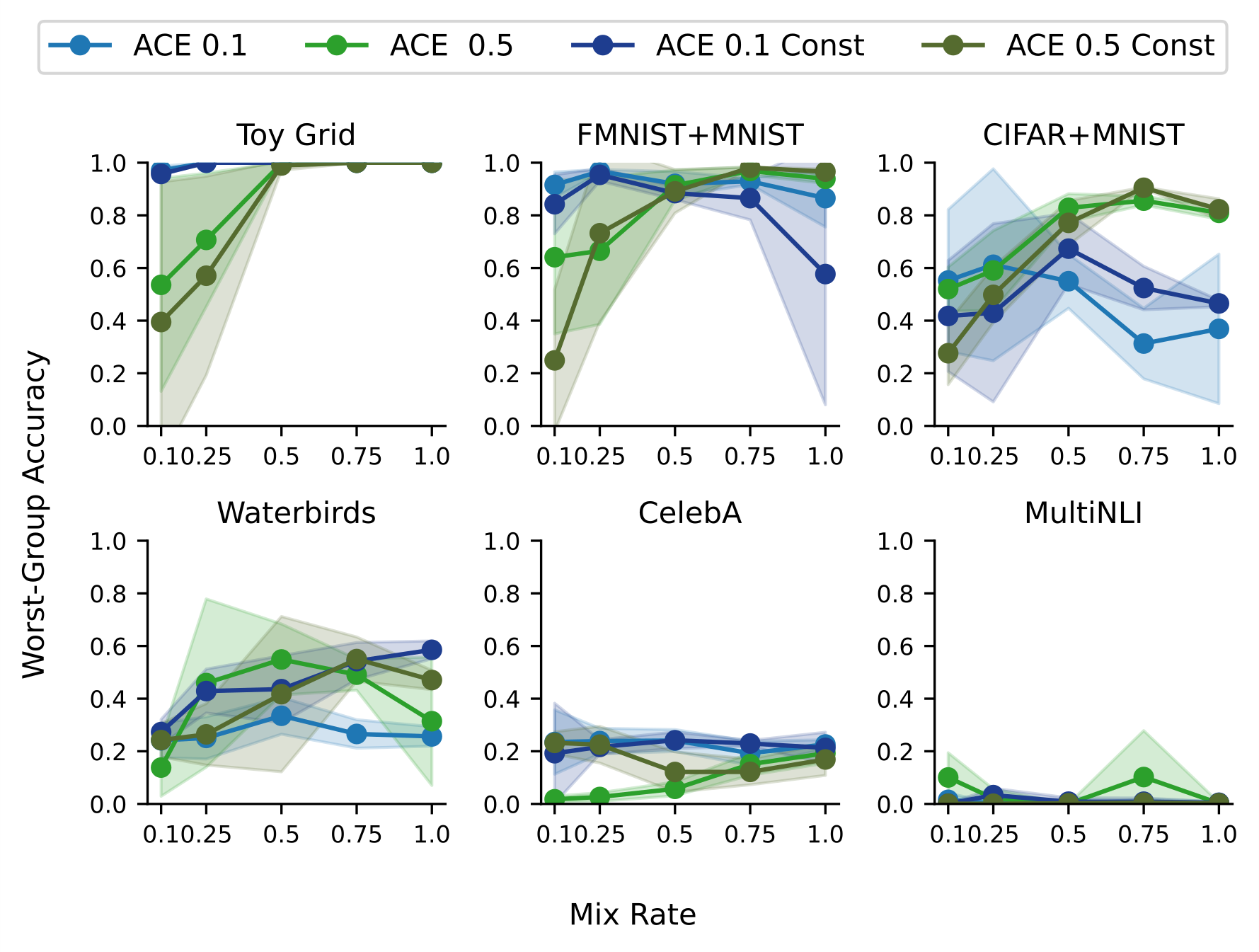}
    \end{subfigure}

    \caption{
        \textbf{(Scheduling Ablation) Accuracy Metrics Across Mix Rates}
    }
    \label{fig:cc_mix_rate_ablation}
\end{figure}

In Figures \ref{fig:cc_mix_rate_avg_ablat_acc_ablation} and \ref{fig:cc_mix_rate_ablation},
we compare performance of ACE TopK with and without mix rate scheduling. 
Mix rate scheduling substantially improves the performance of TopK 0.5 on low mix rates, 
and provides only a modest improvement to TopK 0.1. 

\subsection{Combined Group Loss Ablation}
\label{appendix:comb-group-loss-ablation}
\begin{figure}[ht]
    \centering
    \begin{subfigure}[b]{0.7\textwidth}
        \centering
        \includegraphics[width=\textwidth]{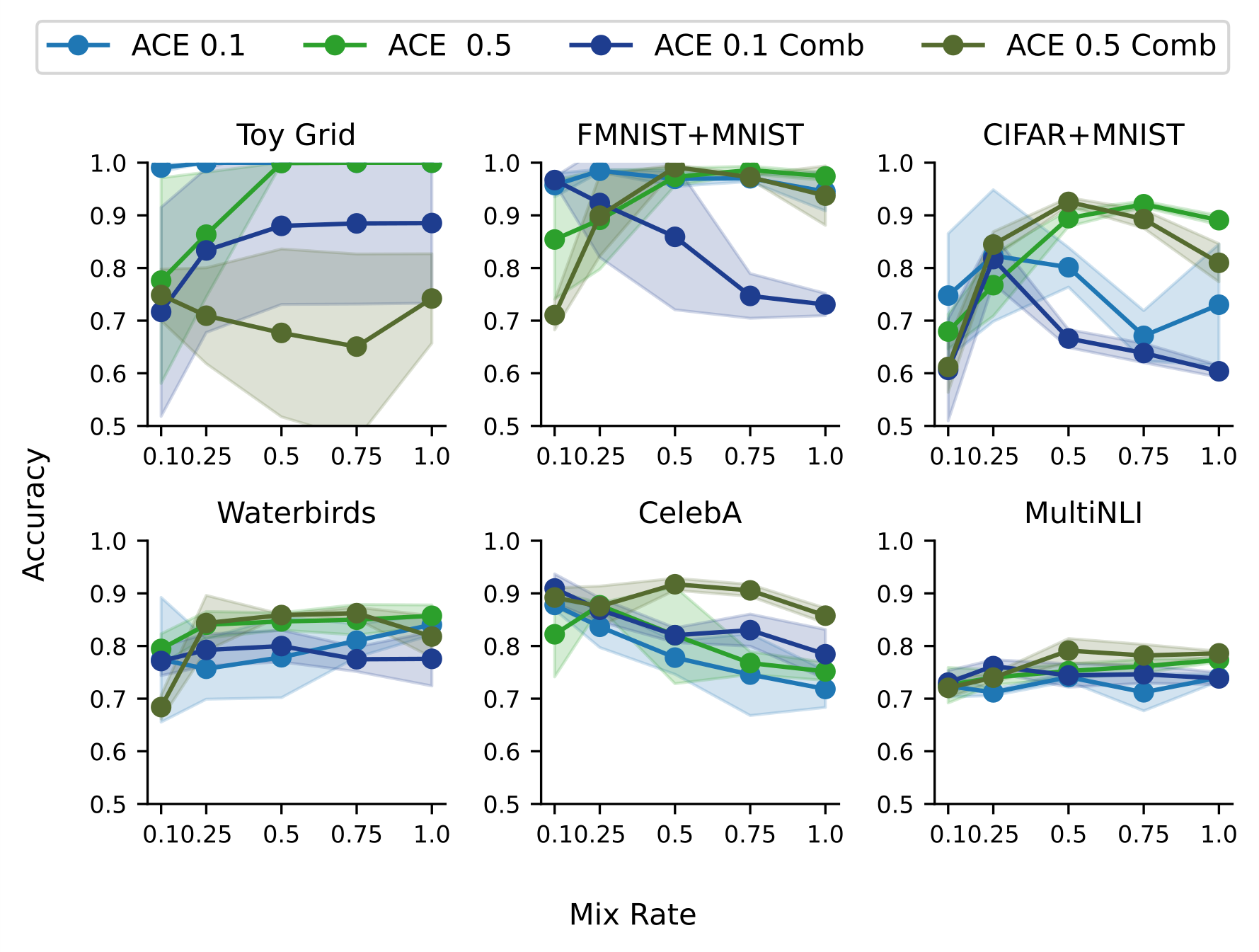}
    \end{subfigure}

    \begin{subfigure}[b]{0.7\textwidth}
        \centering
        \includegraphics[width=\textwidth]{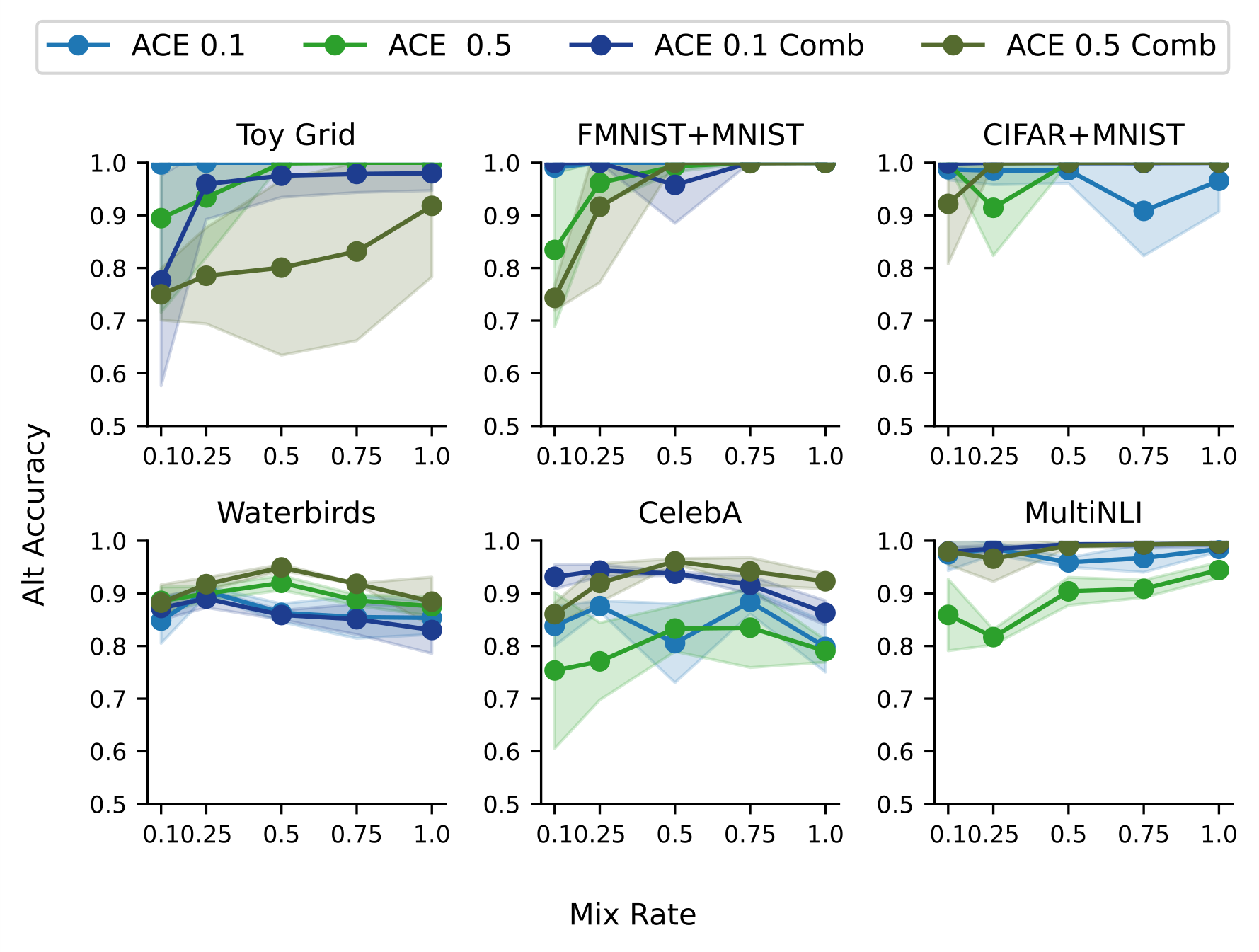}
    \end{subfigure}

    \begin{subfigure}[b]{0.7\textwidth}
        \centering
        \includegraphics[width=\textwidth]{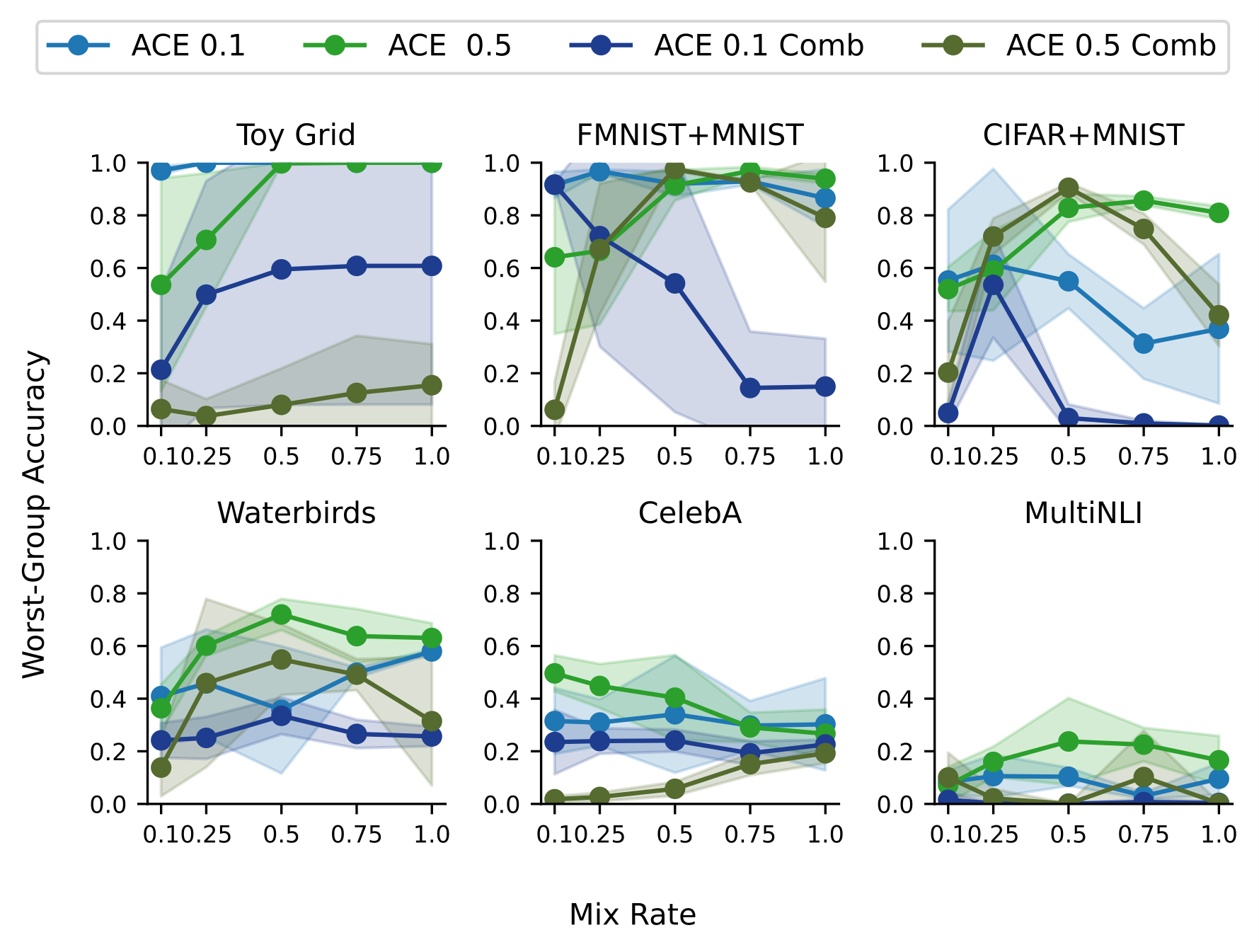}
    \end{subfigure}

    \caption{
        \textbf{(Combined Group Loss Ablation) Accuracy Metrics Across Mix Rates} The Per-group loss lower bounds perform better on class-balanced datasets (top row) while the group-aggregated loss performs better on class-in-balanced datasets (bottom row)
    }
    \label{fig:cc_mix_rate_agg_ablation}
\end{figure}

In the main body, we present ACE results uses separate group losses for evenly-balanced datasets
(Toy Grid, FMNIST-MNIST, and CIFAR-MNIST) and a combined group loss for unbalanced datasets 
(Waterbirds, CelebA, and MultiNLI). In Figure~\ref{fig:cc_mix_rate_agg_ablation},
we compare the performance of separate and combined group losses on each of the six datasets.
The choice of combining or separating group losses has a significant impact on performance.


\section{Model Selection and Test Accuracy Correlations}
\label{appendix:model_selection}

Model selection approaches for early stopping and hyperparameter selection 
typically rely on validation loss or accuracy. But in domain adapation, we 
only have access to source distribution labels, and assume that the target 
distribution will be different in some important way that makes source distribution performance 
a poor proxy for target distribution performance. Prior methods in diverse generalization
(and other domain adaptation work) rely on target labels for model selection and hyperparameter tuning, violating the assumption 
of no labeled access \citep{gulrajani2020searchlostdomaingeneralization}. 

DivDis \citep{lee2023diversifydisambiguatelearningunderspecified} argues that source 
accuracy or loss can be a good proxy, but in the case of complete spurious correlation 
we expect this relationship to break down, as directly learning the spurious feature 
achieves optimal performance. 

\begin{figure}[th]
    \centering
    \includegraphics[width=0.75\textwidth]{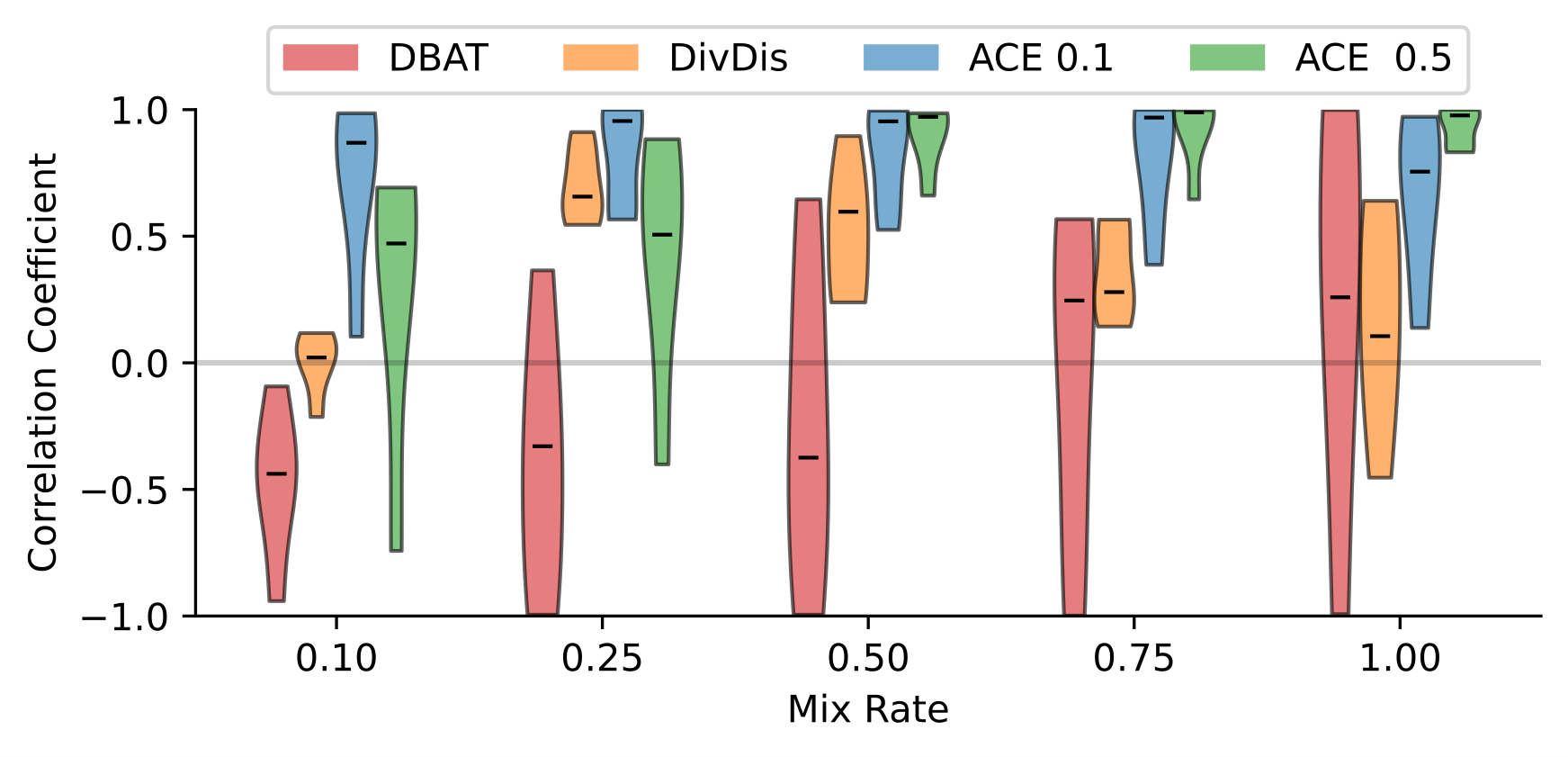}
    \caption{
        \textbf{Correlation Coefficient Between Validation Loss and Test Error}
        Violin plots of the correlation coefficient between validation loss and test error, 
        averaged over 5 datasets and 3 random seeds for each method (we exclude toy grid 
        because we do not use early stopping for this dataset). We find that for all methods, 
        validation loss is a good proxy for test error when the implicit (or explicit)
        assumptions of method match the target distribution mix rate, but is especially 
        strong for ACE. 
    }
    \label{fig:cc_mix_rate_corr_coef_violin}
\end{figure}

\begin{figure}[ht]
    \centering
    \includegraphics[width=0.75\textwidth]{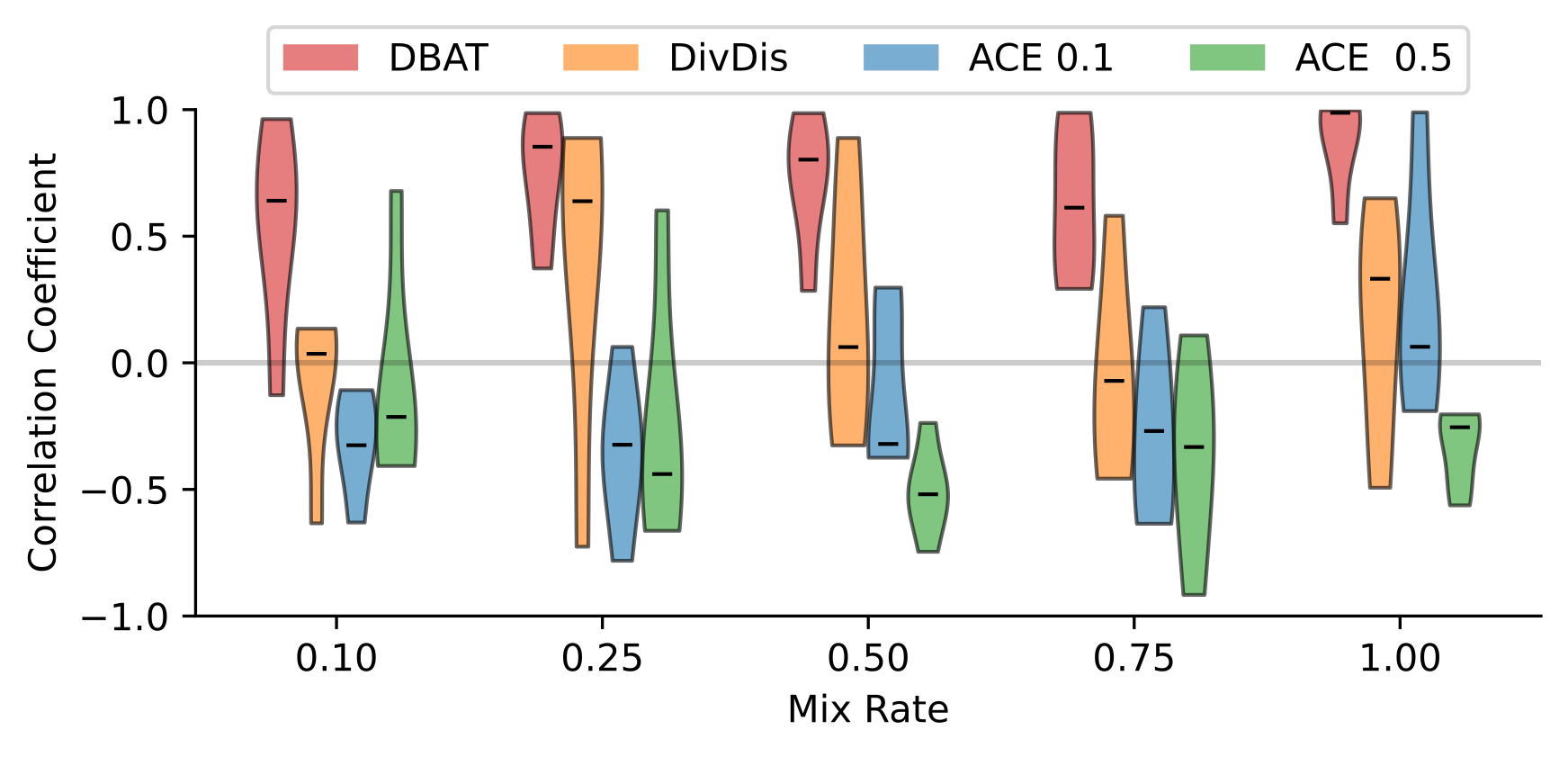}
    \caption{
        \textbf{Correlation Coefficient Between Source Validation Loss and Test Error}
        Violin plots of the correlation coefficient between validation loss and test error, 
        averaged over 5 datasets and 3 random seeds for each method (we exclude toy grid 
        because we do not use early stopping for this dataset). We find that for D-BAT, 
        source validation loss is a good proxy for test error when the mix rate, but negatively 
        correlated with test error for ACE, and approxiamtely uncorrelated for DivDis. 
    }
    \label{fig:cc_mix_rate_corr_coef_source_violin}
\end{figure}

\begin{figure}[ht]
    \centering
    \includegraphics[width=1.0\textwidth]{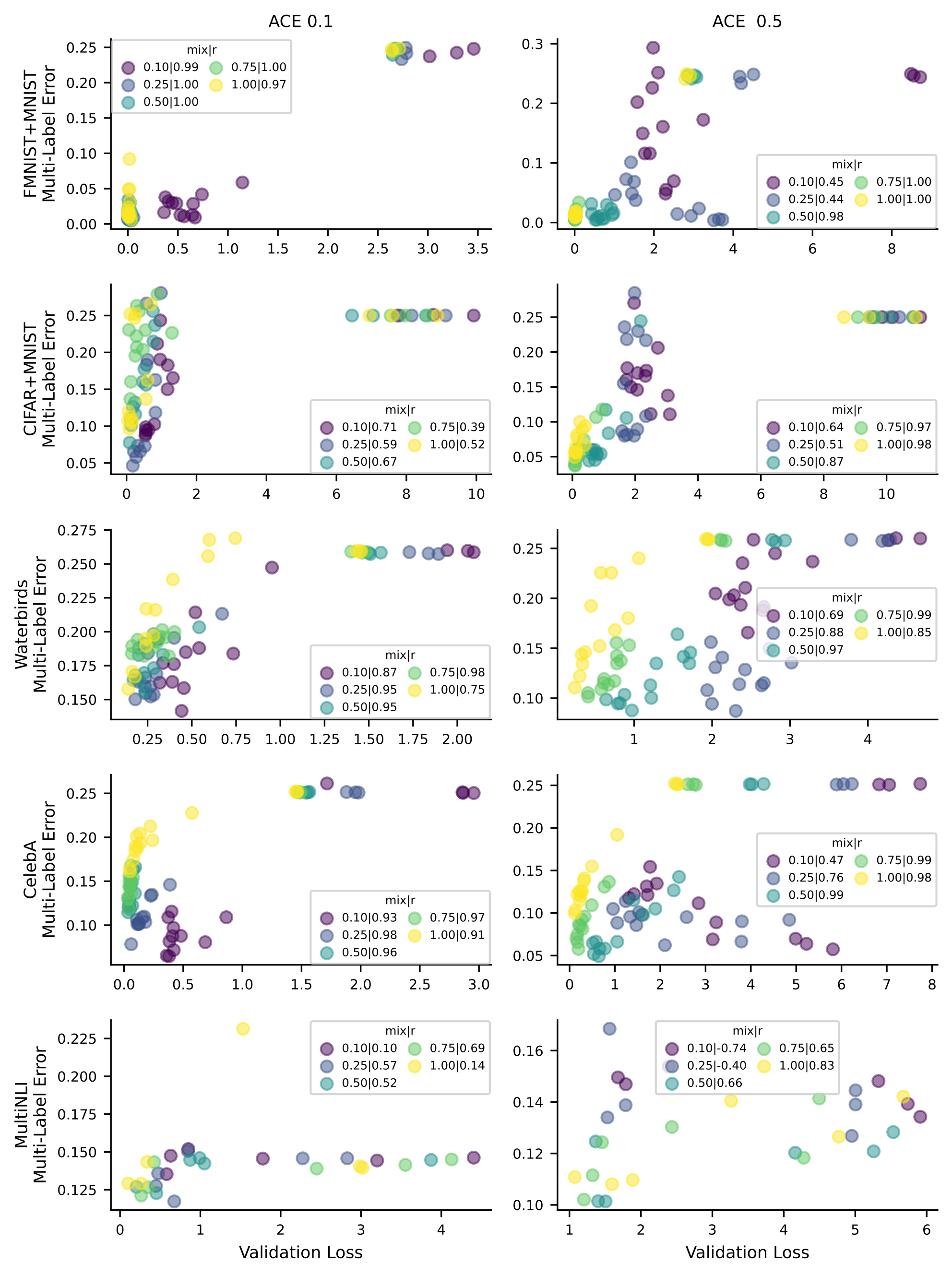}
    \caption{Correlation between Validation Loss and Multi-Label Test Error for ACE TopK}
    \label{fig:val_loss_test_acc_correlation_topk}
\end{figure}

\begin{figure}[ht]
    \centering
    \includegraphics[width=1.0\textwidth]{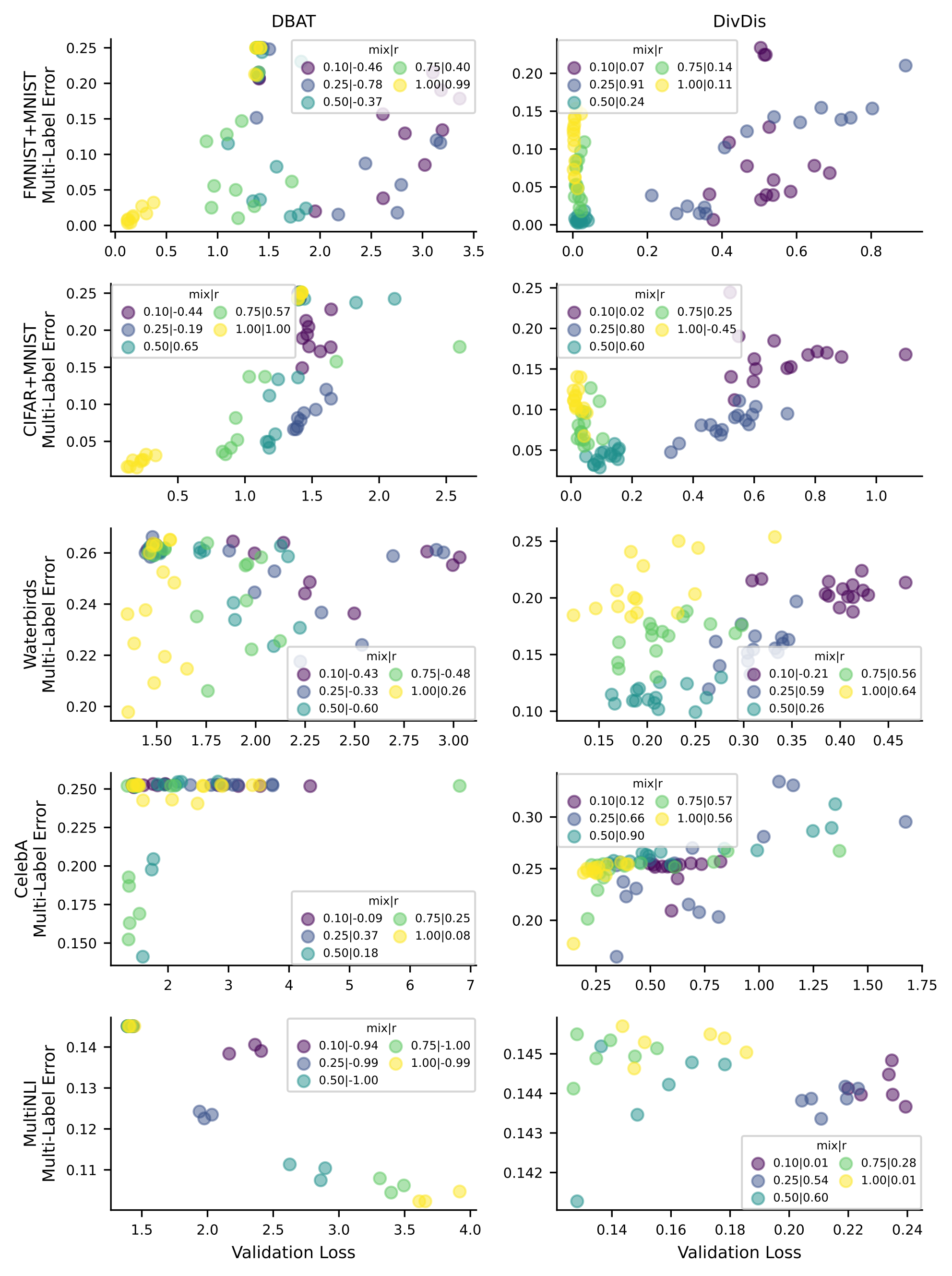}
    \caption{Correlation between Validation Loss and Multi-Label Test Accuracy for DivDis and D-BAT}
    \label{fig:val_loss_test_acc_correlation_divdis_D-BAT}
\end{figure}

\begin{figure}[ht]
    \centering
    \includegraphics[width=1.0\textwidth]{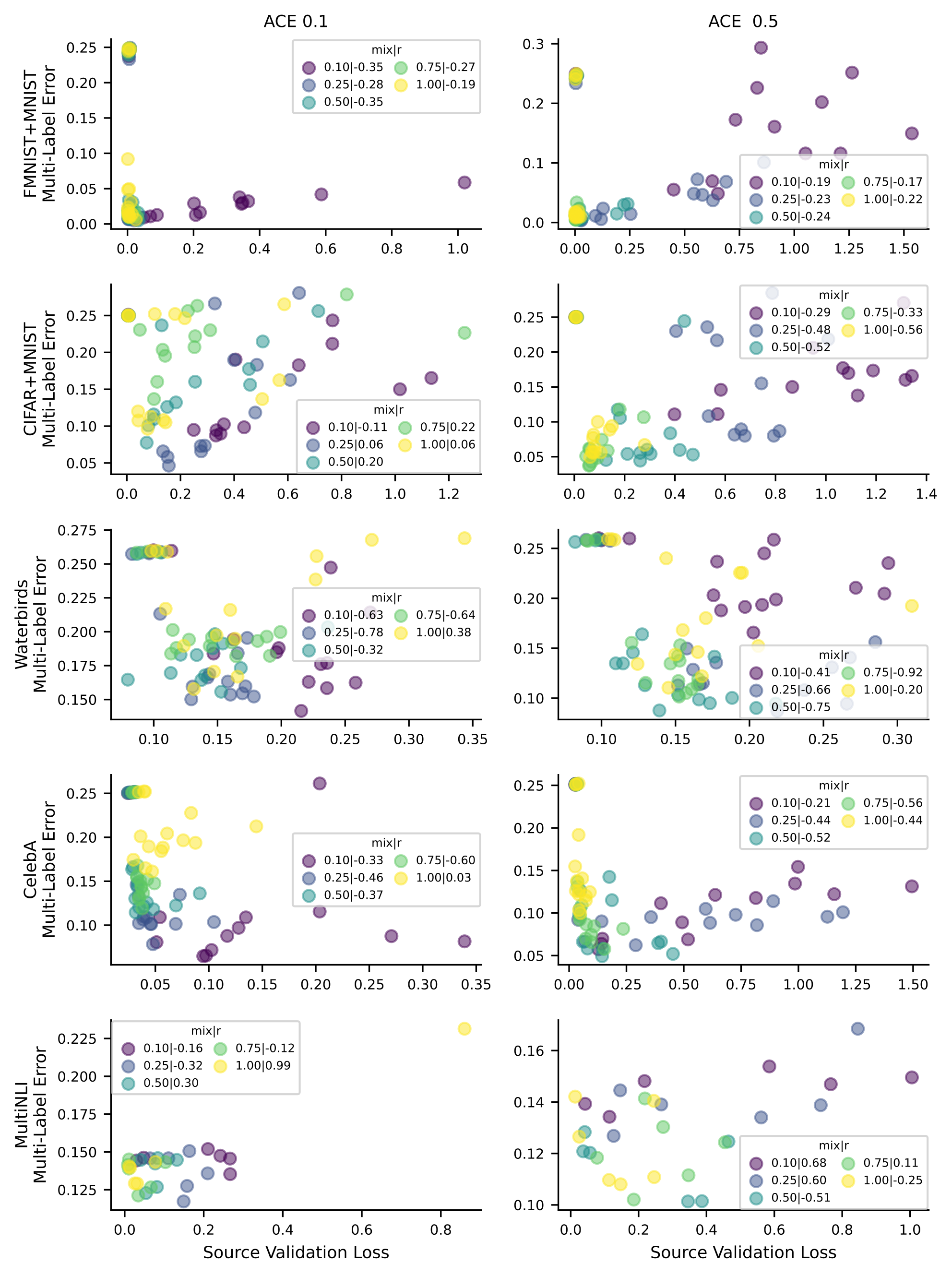}
    \caption{Correlation between Source Validation Loss and Multi-Label Test Error for ACE TopK}
    \label{fig:source_loss_test_acc_correlation_topk_source}
\end{figure}

\begin{figure}[ht]
    \centering
    \includegraphics[width=1.0\textwidth]{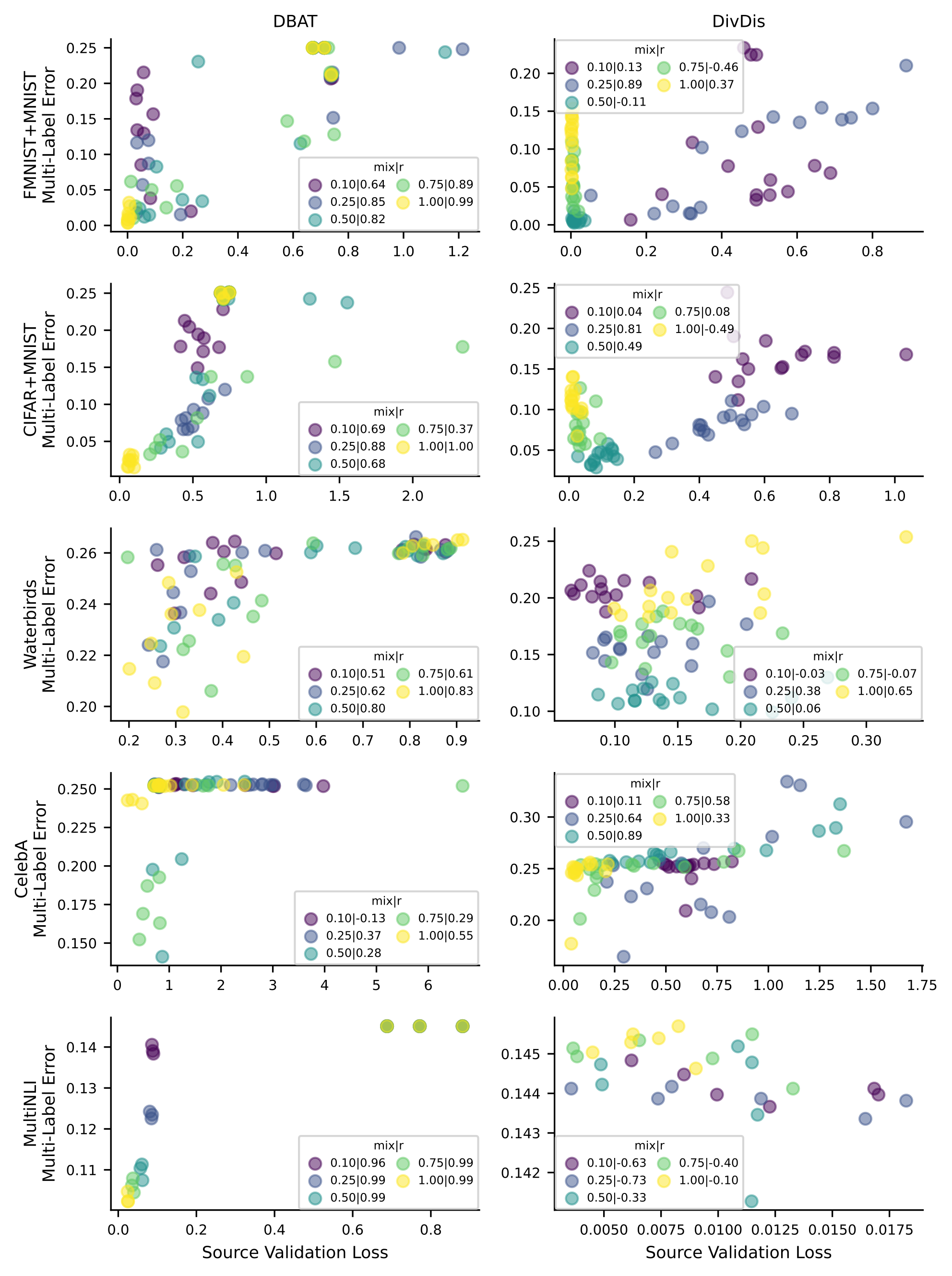}
    \caption{Correlation between Source Validation Loss and Multi-Label Test Error for DivDis and D-BAT}
    \label{fig:source_test_acc_correlation_divdis_D-BAT_source}
\end{figure}

Instead, we use total validation loss (labeled source and unlabeled target) as a proxy 
for target performance \citep{albuqerque2019}. 
This model selection criteria is particularly effective for ACE, 
as the auxillary topk loss corresponds directly to minimizing target distribution 
emperical risk. For DivDis and D-BAT, this model selection method is less principled, 
as we cannot assume an auxillary loss weight of 1 is appropriate. For D-BAT in particular,
where the implicit assumption of the loss function (complete disagreement) rarely holds,
we expect the optimal auxillary loss weight to depend on the mix rate and usually be less 
than one (indeed the original authors find values in the range of 1e-4 to 1e-2 to be effective).
However, without a principled mechanism for tuning this weight, we instead 
treat the DBAT loss function as a kind of constraint, selecting model checkpoints (but 
not hyperparameters) purely from the source source loss. 
See Figures \ref{fig:cc_mix_rate_corr_coef_violin}, 
\ref{fig:val_loss_test_acc_correlation_topk}, \ref{fig:val_loss_test_acc_correlation_divdis_D-BAT} for 
the correlations between total validation loss and test error/accuracy for each method, and 
Figures \ref{fig:cc_mix_rate_corr_coef_source_violin}, \ref{fig:source_loss_test_acc_correlation_topk_source}, 
\ref{fig:source_test_acc_correlation_divdis_D-BAT_source} for the correlations between 
source validation loss and test error/accuracy for each method. 

We find an almost perfect correlation between total validation loss and test accuracy for ACE, 
while for DivDis and D-BAT the strength of the correlation depends on the mix rate. For source validation, 
we find D-BAT's performance is strongly correlated with source loss, 
while the performance of ACE is anti-correlated, and DivDis's performance is uncorrelated. 

\section{Validation Loss Density Estimates}
\label{appendix:val_loss_density_estimates}
\begin{figure}[ht]
    \centering
    \includegraphics[width=0.75\textwidth]{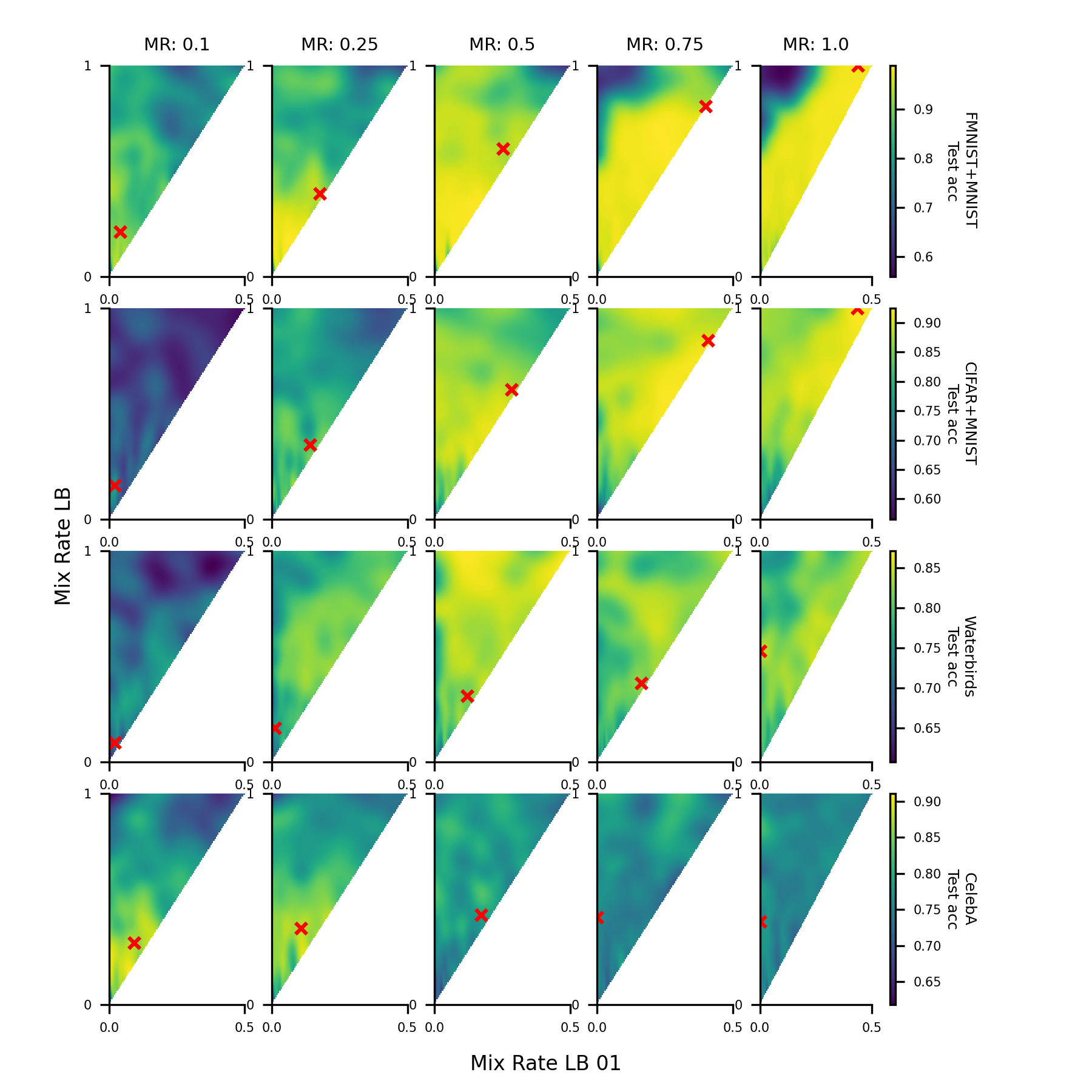}
    \caption{Test Accuracy Density Estimate}
    \label{fig:test_acc_density_grid}
\end{figure}

In Figure \ref{fig:test_acc_density_grid} we present the validation loss density estimates and test accuracy density estimates
for each of the four image datasets. 
We find that the validation loss density estimate provides an especially strong signal for mix rate lower bound
for the synthetic datasets (CIFAR-MNIST, FMNIST-MNIST), and a moderate signal for the real-world datasets 
(Waterbirds-CC, CelebA-CC).

\begin{figure}[ht]
    \centering
    \includegraphics[width=0.75\textwidth]{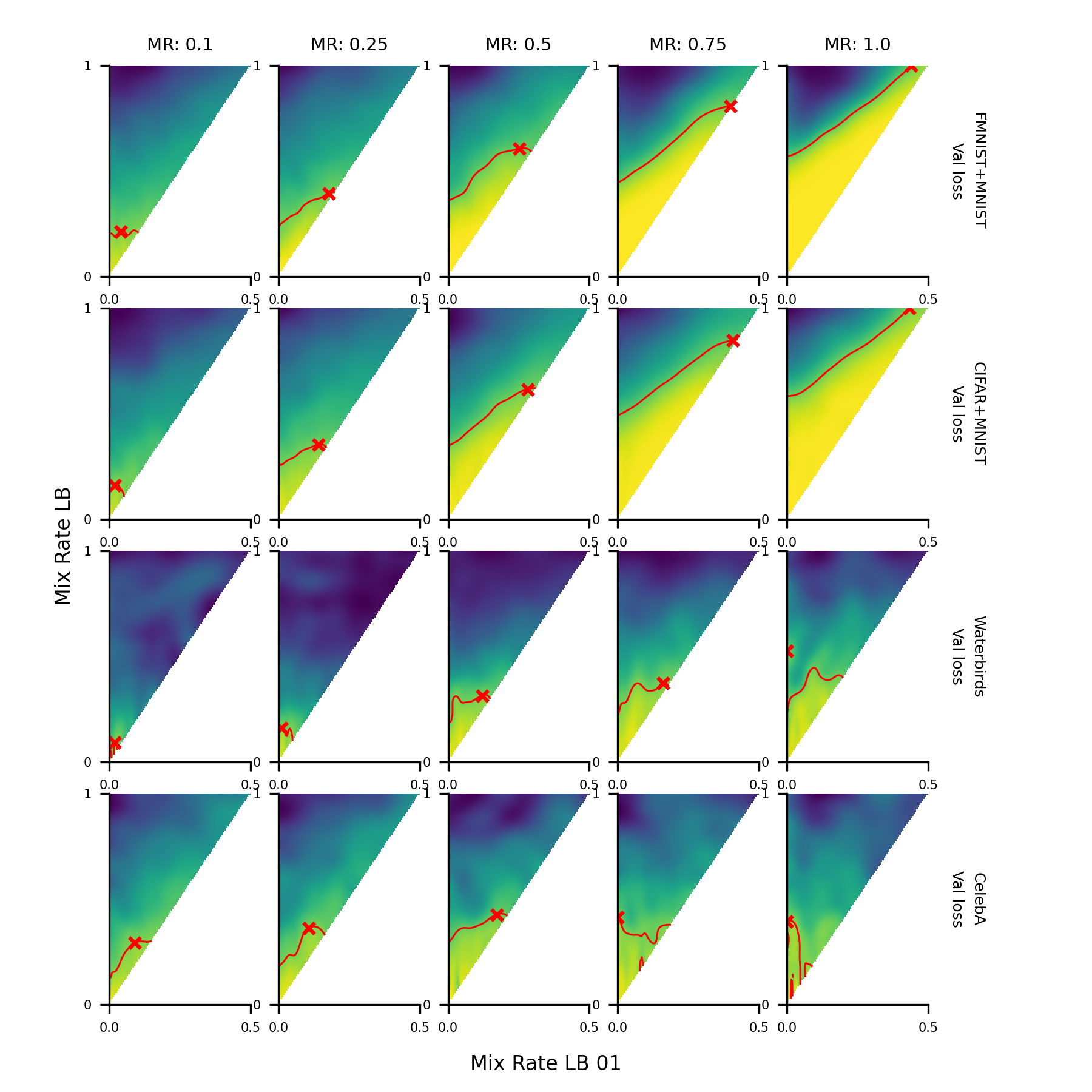}
    \caption{Validation Loss Density Estimate}
    \label{fig:val_loss_density_grid}
\end{figure}

\section{Measurement Tampering Detection Details}
\label{appendix:mtd-details}
We train measurement predictors and tampering detectors on NVIDIA A100's (80GB).

\subsection{Training the Measurement Predictor}
Following \citep{roger2023benchmarksdetectingmeasurementtampering}, we ``pretrain'' 
a measurement predictor with prediction heads for each measurement and an aggregate measurement head 
that predicts the conjunction of all measurements \footnote{
    We train these measurement predictors in a separate repository which will be made available pending acceptance 
}

\subsection{Other Methods}
Here we describe the three measurement tampering detections we use 
to compare ACE and DivDis against. See \citep{roger2023benchmarksdetectingmeasurementtampering} 
for more details.

\textbf{Finetuning on Trusted Distribution (Baseline)} Train a probe (on the aggregate measurement token) 
to distinguish positive and negative instances on the trusted distribution. 

\textbf{Aggregate Measurement Prediction (Baseline)} Use the aggregate measurement prediction head 
from the prediction model (trained on all measurements) to predict the conjunction of all measurements. 

\textbf{Probing for Evidence of Tamper (State of the Art)} Train a probe to distinguish instances where the measurements 
do not agree on the untrusted distribution from all instances on the trusted distribution.

\subsection{Gradient Accumulation for Batch-Dependent Losses}
To compute batch-size dependent losses with gradient accumulation, we first compute the output of each model without gradients, 
then iterate through each micro batch, recomputing parts of the batch with gradients, substituting these outputs for the 
corresponding outputs from the gradient free pass, and computing the loss. 

%

\end{document}